\documentclass[twoside]{article}

\usepackage[preprint]{aistats2027}
\usepackage{commands}

\newcommand{\sweepAnnotations}{Lines and bands are means $\pm1$ standard error over $20$ seeds. Dashed connectors mark configurations at which the seed-paired one-sided Wilcoxon signed-rank test rejects at $p<0.05$; arrows indicate the direction of improvement.}

\begin{document}

%

%

\twocolumn[

  \aistatstitle{VampPrior Unleashed}

  \aistatsauthor{
    Andrew A. Stirn \And
    Abhejit Rajagopal \And
    David A. Knowles \And
    Grace Huynh
  }

  \aistatsaddress{
    Allen Institute \And
    Allen Institute \And
    New York Genome Center \&\\ Columbia University \And
    Allen Institute
  }

]

\begin{abstract}
  Deep latent variable models (DLVMs) underpin many modern analysis pipelines, for example in single-cell genomics, yet their prior distributions have received markedly little attention.
  The VampPrior approximates the optimal prior (the aggregate posterior) and often attains high model evidence.
  Here, we identify two key limitations of the VampPrior: it ties prior and posterior uncertainty to a single amortized covariance and clusters poorly due to its fixed uniform mixture weights.
  We \emph{unleash} the VampPrior by decoupling its prior and posterior covariances, resolving the conflict of generating high prior and low posterior entropy with the same shared parameters and substantially improving model evidence.
  We then replace the uniform mixture with a Dirichlet process.
  When combined, our proposals provide higher model evidence than the VampPrior and robust non-parametric clustering.
  Augmenting single cell Variational Inference (scVI) with our prior improves single-cell integration and clustering across six atlases while significantly raising model evidence.
  Our approach changes only the prior and thus can be applied to any DLVM with a continuous latent space.
\end{abstract}

\section{Introduction}

Deep latent variable models (DLVMs) underpin many of the popular single-cell models of the last decade: scVI \parencite{lopezDeepGenerativeModeling2018} and scANVI \parencite{xuProbabilisticHarmonizationAnnotation2021} for scRNA-seq; PeakVI \parencite{ashuachPeakVIDeepGenerative2022} and PoissonVI \parencite{martensModelingFragmentCounts2024} for ATAC-seq; MethylVI and MethylANVI \parencite{weinbergerProbabilisticModellingSinglecell2026} for scBS-seq; totalVI \parencite{gayosoJointProbabilisticModeling2021} for Cite-seq; MultiVI \parencite{ashuachMultiVIDeepGenerative2023} for single-cell multi-omics; and DestVI \parencite{lopezDestVIIdentifiesContinuums2022}, ResolVI \parencite{ergenResolVIAddressingNoise2025}, and scVIVA \parencite{levyScVIVAProbabilisticFramework2025} for spatial transcriptomics.
Single-cell analyses benefit from explicit probabilistic modeling to handle sparse, noisy, and over-dispersed count data; batch effects (nuisance technical variation); and partially labeled data.
DLVMs provide an integrated low-dimensional, batch-effect-free representation of the data that enables downstream tasks such as clustering, visualization, trajectory inference, and differential expression analysis.
Despite widespread adoption of DLVMs by biologists, there has been little innovation in the prior distributions used in these models.
Prior choice governs the geometry of the latent space, affecting posterior collapse, representation quality, and, critically for single-cell applications, the recovery of discrete cell-type clusters and continuous differentiation trajectories.
Roughly half of the aforementioned methods employ a standard normal prior, while the other half employ a Gaussian mixture model (GMM) prior.
DestVI is the only method to leverage the VampPrior \parencite{tomczakVAEVampPrior2018} (albeit alongside a standard normal prior).

Observing an awkward tradeoff between a GMM prior's clustering quality and the VampPrior's model evidence, \textcite{stirnVampPriorMixtureModel2025} introduced the VampPrior Mixture Model (VMM).
They tune the VMM to achieve state-of-the-art clustering but find there its model evidence lags behind the VampPrior's, suggesting the model is a less faithful descriptor of the observed data.
This paper asks, ``Can we have the best of both worlds?'' and ``How can we easily traverse the tradeoff between structure and model evidence?''
We propose a new nonparametric prior that definitively shifts the Pareto frontier of the clustering-quality and model-evidence tradeoff.
At their most-balanced operating points, our approach outperforms the VMM at both clustering and model evidence.

Our technical contributions are twofold.
First, we recognize the VampPrior's decision to model prior and posterior variance with the same shared parameters hamstrings its modeling performance since the same covariance parameterizing function has the conflicting tasks of producing both low entropy posteriors that drive predictive performance and a higher entropy prior that governs the marginal likelihood.
Our \emph{VampPrior-Unleashed} frees the VampPrior from this artificial constraint by decoupling the VampPrior's component covariances from the encoder network.
We find that this additional degree of freedom greatly improves model evidence.
Second, we replace the VampPrior's uniform mixture with a Dirichlet process (DP) mixture to encourage fewer active components.
Empirically, we observe that a DP VampPrior (without decoupled covariances) exacerbates evidence degradation making our proposed covariance decoupling even more valuable if using a DP.
In contrast, at its its most-balanced operating point our \emph{DP-VampPrior-Unleashed} (a DP-mixture VampPrior with decoupled covariances) yields a well-structured latent distribution with cohesive, parsimonious clusters \emph{and} significantly higher evidence than even the VampPrior's highest attainable---a substantial achievement given the VampPrior already approximates the optimal prior for the evidence lower bound (ELBO).
Conveniently, adjusting the DP's concentration traverses the Pareto frontier without needing to specify the number of mixture components a priori.

Augmenting scVI with our \emph{DP-VampPrior-Unleashed} prior improves its single-cell integration benchmark (scIB) \textcite{lueckenBenchmarkingAtlaslevelData2022} metrics while attaining state-of-the-art levels of marginal likelihood enabling higher-fidelity downstream differential expression analyses.
We emphasize that our contributions apply to any continuous latent-space DLVM despite our experimental focus on single-cell data.
We anticipate many more applications may benefit from our approach in biology and beyond.

\section{Background}

\subsection{Deep Latent Variable Models}

DLVMs are a class of generative models that explicitly parameterize the likelihood of observed variables with neural networks operating on latent variables.
The variational autoencoder (VAE) \parencite{kingmaAutoEncodingVariationalBayes2014,rezendeStochasticBackpropagationApproximate2014} is a DLVM that provides computationally efficient variational inference (VI) of low-dimensional, latent Gaussian variables $Z$ that explain observed high-dimensional data $X$.
\textcite{kingmaAutoEncodingVariationalBayes2014} choose $p(Z)\triangleq\N(0,I)$ as the prior, some suitable likelihood $p_{\theta}(X|Z=z)\triangleq p(X;f(z;\theta))$ (e.g., Bernoulli or Normal) parameterized by neural network $f$ with parameters $\theta$ operating on $z$, and posit amortized variational family $q_{\phi}(Z|X=x)\triangleq \N(Z;g(x;\phi))$ (encoding distribution), where neural network $g$ with parameters $\phi$ maps $x$ onto the posterior parameter space.
They then fit $\theta$ and $\phi$ using black box VI~\parencite{ranganathBlackBoxVariational2014} with reparameterization gradients~\parencite{williamsSimpleStatisticalGradientfollowing1992} to maximize the evidence lower bound (ELBO) on the marginal likelihood of the data,
\begin{align}
  \log p(\D) \geq \sum_{x\in\D} & \E_{q_{\phi}(Z|X=x)}\big[\log p_{\theta}(X=x|Z) \big] - \nonumber
  \\ &\KL\big(q_{\phi}(Z|X=x)\ \|\ p(Z)\big). \label{eq:vae-elbo}
\end{align}

\subsection{The VampPrior}

\textcite{hoffmanElboSurgeryAnother2016,tomczakVAEVampPrior2018} identify the prior that maximizes \cref{eq:vae-elbo} as the aggregate posterior,
\begin{align}
  \label{eq:aggregate-posterior}
  p^{*}(Z)=\frac{1}{N}\sum_{j=1}^{N}q_{\phi}(Z|X=x_j),
\end{align}
taken over the $N$ training points.
Motivated by Empirical Bayes, \textcite{tomczakVAEVampPrior2018} approximate $p^{*}(Z)$ using $K \ll N$ randomly selected (without replacement) training data points (pseudo-inputs) instead of all $N$ for computational tractability.
They call their approximation to the Empirical Bayes solution the \emph{VampPrior} and define it as a uniform mixture of $K$ variational posteriors,
\begin{align}
  \label{eq:vampprior}
  p(Z) \triangleq \frac{1}{K}\sum_{j=1}^K q_{\phi}(Z|X=u_j)
\end{align}
To train a VAE with a VampPrior, one sets $p(Z)$ according to \cref{eq:vampprior} and maximizes \cref{eq:vae-elbo} w.r.t. $\theta$, $\phi$, and $\{u_j\}_{j=1}^K$---training the pseudo-inputs $u$ (prior parameters) alongside the encoder and decoder parameters.
Despite our observed weakness in its approximation, the VampPrior often achieves higher marginal likelihoods than other priors \parencite{tomczakVAEVampPrior2018,stirnVampPriorMixtureModel2025}.

\subsection{Dirichlet Process Mixtures}

A Dirichlet Process (DP), $\DP(\alpha, G_0)$, is a stochastic process whose realizations are themselves probability distributions.
A positive concentration parameter $\alpha$ and a base distribution $G_0$ parameterize a DP.
A realization $G \sim \DP(\alpha, G_0)$ is almost surely discrete taking the form $G = \sum_{k=1}^{\infty} \pi_k \delta_{\theta_k}$, where $\delta_{\theta_k}$ is a point mass at $\theta_k$, itself drawn independently from $G_0$.
Sethuraman’s stick-breaking construction \parencite{sethuramanConstructiveDefinitionDirichlet1994} explicitly characterizes the construction of the infinite sequence of mixture weights, $\pi = (\pi_1, \pi_2, \dots)$: after sampling the stick-breaking proportions $v_k \sim \Bet(1, \alpha)$ one sets $\pi_k \triangleq v_k \prod_{j=1}^{k-1} (1 - v_j)$.
Informally, starting with an initial stick of length one, each iterative break takes a $v_k$ proportion of what remains.
The GEM (Griffiths-Engen-McCloskey) distribution \parencite{griffithsLinesDescentDiffusion1980,engenNoteGeometricSeries1975,mccloskeyModelDistributionIndividuals1965} characterizes the joint distribution $\pi \sim \text{GEM}(\alpha)$ and ensures that the infinite sequence of weights sums to one almost surely and naturally encourages a sparse, size-biased allocation of probability mass across latent clusters.
For this reason, a DP is a popular choice in Bayesian nonparametrics when the number of mixture components is unknown a priori.

\section{Methods}
\label{sec:methods}

\subsection{Unleashing the VampPrior}
\label{sec:methods:unleashing}

To the best of our knowledge, we are the first to recognize that the VampPrior's choice to model both prior and posterior variance with the same shared parameters $\phi$ results in conflict between the desiderata of low entropy posteriors \emph{and} a high entropy prior.
We demonstrate empirically that this constricted construction is suboptimal from a marginal likelihood perspective and therefore propose decoupling the VampPrior's component variances from the encoder network to resolve this conflict.
Concretely, we propose replacing \cref{eq:vampprior} with,
\begin{align}
  \label{eq:vampprior-decoupled}
  p(Z) \triangleq \frac{1}{K}\sum_{k=1}^K \N\Big(\E_{q_{\phi}}[Z|X=u_k], \Sigma_j\Big).
\end{align}
Here, $\E_{q_{\phi}}[Z|X=u_k]$ denotes the encoder's mean output when operating on pseudo-input $u_k$, and $\Sigma_k$ is the $k$-th of $K$ trainable covariance matrices.
With $\Sigma_k$ exclusively a prior parameter, the prior is free to increase entropy as needed to approximate the Empirical Bayes solution (\cref{eq:aggregate-posterior}) leaving $q_{\phi}(Z|X=x)$'s covariance free to reduce entropy and drive predictive performance.
We show this decoupling substantially improves evidence relative to the original VampPrior.

While the covariance is decoupled, the amortized mean function, $\E_{q_{\phi}}[Z|X=u_k]$, remains shared between prior and posterior.
When initializing pseudo-inputs $u$ by sampling from the training data, the decoupled VampPrior retains the VampPrior's property of well-initialized prior components.
\Cref{eq:vampprior-decoupled} contains the original VampPrior as a special case when $\Sigma_k = \Cov_{q_{\phi}}[Z|X=u_k]$ for all $k$.

\subsection{A Dirichlet Process VampPrior}
\label{sec:methods:dp}

\textcite{tomczakVAEVampPrior2018} observe that increasing $K$ increases the ELBO of a VAE with a VampPrior; this makes sense since increasing the entropy of the VampPrior requires either increasing the component entropy or increasing the number of components.
In other words, increasing $K$ reduces the burden on the encoder to produce high entropy priors while maintaining low entropy posteriors.
Indeed, in the limit $K \rightarrow N$ there is no prior-posterior conflict.
Reducing $K$ exacerbates this burden. 
Often one uses a large $K$ (several hundred) to better approximate the aggregate posterior, but it is this regime where the VampPrior's cluster assignments become less meaningful.
With the additional degree of freedom afforded by our proposal to decouple covariance, we can now effectively reduce the number of prior components by replacing the uniform mixture in \cref{eq:vampprior,eq:vampprior-decoupled} with a DP mixture using a stick-breaking construction,
\begin{align}
  v_1,\hdots,v_{\infty} & \sim \Bet(1, \alpha) \label{eq:stick-breaks}         \\
  \pi_k                 & = v_k\prod_{j=1}^{k-1}(1-v_j) \label{eq:proportions} \\
  c_i\mid\Pi =\pi       & \sim \Cat(\pi) \label{eq:assignment}                 \\
  z_i\mid C_i=c_i       & \sim p(Z \mid C_i = c_i) \label{eq:component}        \\
  x_i\mid Z_i = z_i     & \sim p_{\theta}(X  \mid Z_i = z_i). \nonumber
\end{align}
Here, $p(Z \mid C_i = k)$ denotes the prior's $k$-th component distribution, which can be either $q_{\phi}(Z|X=u_k)$ for the original VampPrior (\cref{eq:vampprior}) or $\N(\E_{q_{\phi}}[Z|X=u_k], \Sigma_k)$ for the VampPrior with decoupled covariance (\cref{eq:vampprior-decoupled}).
Sampling the cluster assignment $c_i$ allows us to index a prior component distribution rather than marginalizing over all components as \cref{eq:vampprior,eq:vampprior-decoupled} have to do without an expanded model.

Following~\textcite{bleiVariationalMethodsDirichlet2004}, we approximate the generative DP with a truncated variational family, defining the approximate posterior on stick-breaking proportions $q(V_j) \triangleq \Bet(\gamma_{j,1}, \gamma_{j,2})$ for $j=1,\hdots,K-1$
with variational parameters $\gamma$.
For $k<K$, (sampled) posterior mixture proportions follow \cref{eq:proportions}.
The final break ($k=K$) deterministically takes the remainder of the stick with $\pi_K  \triangleq 1 - \sum_{j=1}^{K-1} \pi_j$.
When estimating the the log marginal likelihood of the data with an Importance Weighted Autoencoder (IWAE) estimator \parencite{burdaImportanceWeightedAutoencoders2016}, this deterministic construction of $\pi_K$ avoids the mathematical issues associated with a Dirac delta proposal distribution.
We optimally update $q(C_i|X_i=x_i) \varpropto $
\begin{align*}
  \exp\bigg(\E_{q_{\phi}(Z|X_i=x_i)} \Big[\log p(Z|C_i=j)\Big] + \E_{q(V)}\Big[\log \Pi_j\Big]\bigg),
\end{align*}
where the expectation of $\E_{q(V)}[\log \Pi_j]$ integrates analytically after substituting $\Pi$'s stick-breaking construction from $V$ (\cref{eq:proportions}).
With all priors and variational families defined we can write down the VAE's ELBO for a VampPrior with a DP mixture as:
\begin{align}
  \label{eq:elbo-dp}
   & \Bigg(\sum_{i=1}^N \E_{q_\phi(Z|X=x_i)}\Big[\log p_{\theta}(X=x_i|Z)\Big]\Bigg) - \nonumber                     \\
   & \Bigg(\sum_{i=1}^N \E_{q(C_i|X_i=x_i)}\Big[\KL\big(q_{\phi}(Z|X=x_i)\ \|\ p(Z|C_i)\big)\Big]\Bigg) -  \nonumber \\
   & \Bigg(\sum_{i=1}^N \E_{q(V)}\Big[\KL\big(q(C_i|X_i=x_i)\ \|\ p(C_i|V)\big)\Big] \Bigg) - \nonumber              \\
   & \Bigg( \sum_{j=1}^{K-1} \KL\big(q(V_j;\gamma_{j,1},\gamma_{j,2})\ \|\ p(V_j;1,\alpha])\big) \Bigg)
\end{align}
We jointly optimize \cref{eq:elbo-dp} w.r.t.\ $\theta$, $\phi$, $\{u_j\}_{j=1}^K$, $\gamma$, and, in the case of decoupled covariance, $\{\Sigma_j\}_{j=1}^K$.

Using a truncated mixture for the variational family only requires defining $K$ prior components, enabling us to use either the VampPrior's components in \cref{eq:vampprior} or our decoupled components in \cref{eq:vampprior-decoupled}.
Specifically, we can set $p(Z|C=c)$ (\cref{eq:component}) to $q_{\phi}(Z|X=u_c)$ for the original VampPrior or to $\N(\E_{q_{\phi}}[Z|X=u_j], \Sigma_c)$ for our VampPrior-Unleashed.

\subsection{Related Work}
\label{sec:methods:related-work}

\Cref{fig:vae_priors} (\cref{supp:prior-lineage}) situates our contributions within a lineage of VAE priors.
The motivations underlying these developments are generally centered around structured latent representations, flexible prior modeling, or both.
From the VAE's \parencite{kingmaAutoEncodingVariationalBayes2014} standard normal prior, \textcite{jiangVariationalDeepEmbedding2017} replace $p(Z)$ with a parametric GMM prior (VaDE), which \textcite{stirnVampPriorMixtureModel2025} extend to a Bayesian GMM with a Dirichlet prior to flexibly model the number of clusters.
\textcite{tomczakVAEVampPrior2018} approximate the aggregate posterior (\cref{eq:aggregate-posterior}).
The VampPrior Mixture Model (VMM) \parencite{stirnVampPriorMixtureModel2025} unifies both branches by amortizing their Bayesian GMM's component means through the encoder, thereby containing the VampPrior as a special case while retaining an approximate DP mixture.
\textcite{stirnVampPriorMixtureModel2025} report their VMM dominates both VaDE and their Bayesian GMM at both log marginal likelihood and NMI.
We therefore consider the VMM as the best representative of this lineage and exclude their Bayesian GMM and VaDE from our experiments.

\begin{figure*}[t]
  \centering
  \includegraphics[width=\textwidth]{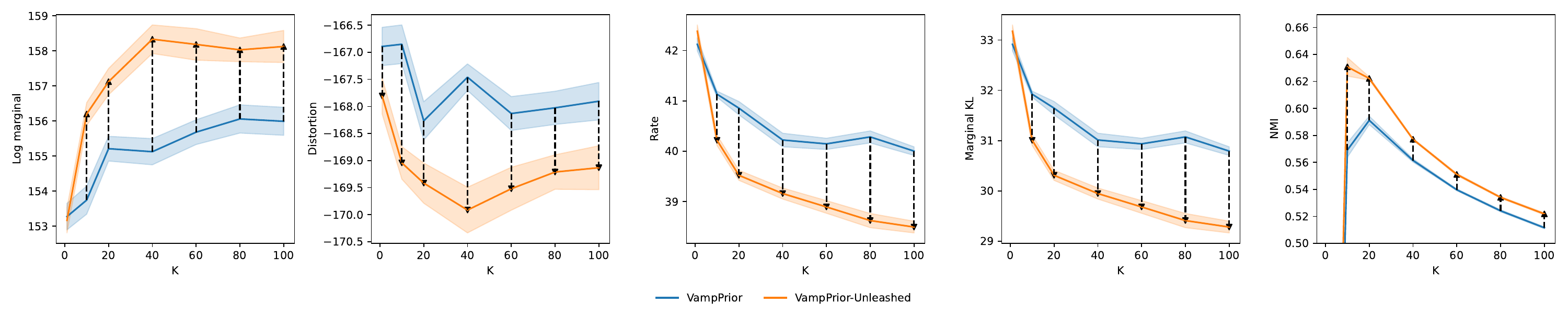}
  \caption{Effect of varying the number of components $K$ on FashionMNIST with $u$ initialized from randomly selected training data. \sweepAnnotations}
  \label{fig:sweep_k:fashion_mnist:data}
\end{figure*}

\section{Experiments}
\label{sec:experiments}

Our experiments investigate how prior selection affects various DLVMs' performances across different datasets and domains.
Throughout, we explore the tradeoff between marginal likelihood and clustering quality.
We estimate the marginal likelihood using the IWAE bound \parencite{burdaImportanceWeightedAutoencoders2016} with $5,000$ samples per data point.
We choose normalized mutual information (NMI) to measure clustering quality.
\Cref{supp:experimental-setup:metrics} elaborates further and includes an IWAE bound for global latent variables.
Commonly, one initializes VampPrior pseudo-inputs with randomly selected training data.
We consider two additional strategies for initializing the pseudo-inputs: noise and stratified sampling via coarse labels, which are often available in scRNA-seq.
\Cref{supp:experimental-setup:initialization} motivates and details these strategies.

\subsection{Unleashing the VampPrior}
\label{sec:experiments:unleashing}

We isolate the effect of decoupling the VampPrior's component covariances from the encoder by comparing the difference in performance of the \textbf{VampPrior} (\cref{eq:vampprior}) and our \textbf{VampPrior-Unleashed} (\cref{eq:vampprior-decoupled}) in a VAE.
\Cref{fig:sweep_k:fashion_mnist:data} sweeps the number of components $K$ on FashionMNIST.
In addition to the per-datum marginal likelihood and NMI (first and last panel), we examine distortion (negative of \cref{eq:vae-elbo}'s first term) and rate (\cref{eq:vae-elbo}'s second term) \parencite{alemiFixingBrokenELBO2018} as well as marginal Kullback--Leibler (KL) divergence \parencite{hoffmanElboSurgeryAnother2016} to distinguish log marginal differences.
\Cref{supp:sweep_k} contains FashionMNIST sweeps for different $u$ initialization strategies (\cref{fig:sweep_k:fashion_mnist}) and the corresponding MNIST sweeps (\cref{fig:sweep_k:mnist}).

The results are unambiguous and diagnostic of the conflict identified in \cref{sec:methods:unleashing}.
Across two datasets, three pseudo-input initialization strategies, and $K\geq10$ ($36$ configurations in total), the decoupled variant attains the better mean on \emph{every} configuration for \emph{every} metric.
Differences are significant in $31/36$ configurations for log marginal, $32/36$ for distortion, and $36/36$ for rate, marginal KL, and NMI.
With \texttt{data} $u$-initialization, log marginal's per-datum gain broadly grows with $K$---peaking at $+2.1$ nats at $K=100$ on FashionMNIST (\cref{fig:sweep_k:fashion_mnist:data}) and $+2.8$ nats at $K=80$ on MNIST (\cref{fig:sweep_k:mnist:data}).
The largest observed gain is $+3.6$ nats at $K=100$ on MNIST under \texttt{random} $u$-initialization (\cref{fig:sweep_k:mnist:random}).
Exponentiating these per-datum differences and multiplying by $60{,}000$ (number of samples) results in decisive Bayes factors \parencite{kassBayesFactors1995} favoring the \textbf{VampPrior-Unleashed}.

The rate and distortion panels explain evidence gains and confirm our core hypothesis.
Decoupling lowers rate for $K\geq10$ by roughly $0.8$--$1.7$ nats.
Freed from inflating $\Cov_{q_\phi}[Z|X=u_j]$ for prior entropy, the encoder emits sharper posteriors that are now cheaper to code under a prior whose components can widen on their own.
Distortion also falls making the decoupled model simultaneously better on both ELBO terms (\cref{eq:vae-elbo}).
This zero-tradeoff behavior indicates a constraint relaxation rather than a repositioned operating point on a fixed rate-distortion curve.
The marginal KL panels captures the cost a chosen prior incurs relative to the optimal prior (\cref{eq:aggregate-posterior}).
While the optimal prior is different for every model since it depends on the final encoder, it is clear decoupling better approximates its own optimal prior.

\subsection{Clustering with the VampPrior}
\label{sec:experiments:clustering}

\begin{figure*}
  \centering
  \begin{subfigure}[t]{0.3\textwidth}
    \centering
    \includegraphics[width=\textwidth]{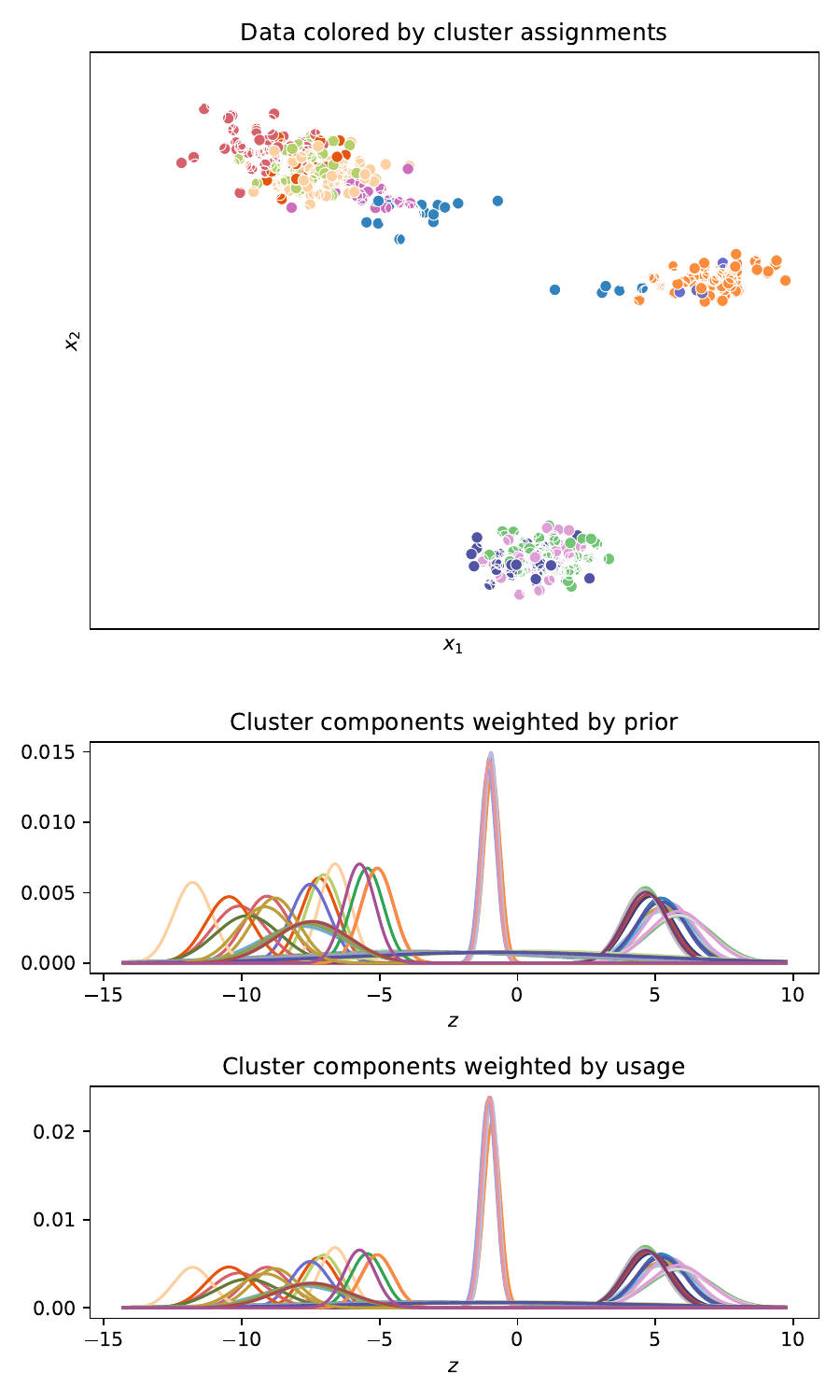}
    \caption{VampPrior}
    \label{fig:toy:original}
  \end{subfigure}
  \hfill
  \begin{subfigure}[t]{0.3\textwidth}
    \centering
    \includegraphics[width=\textwidth]{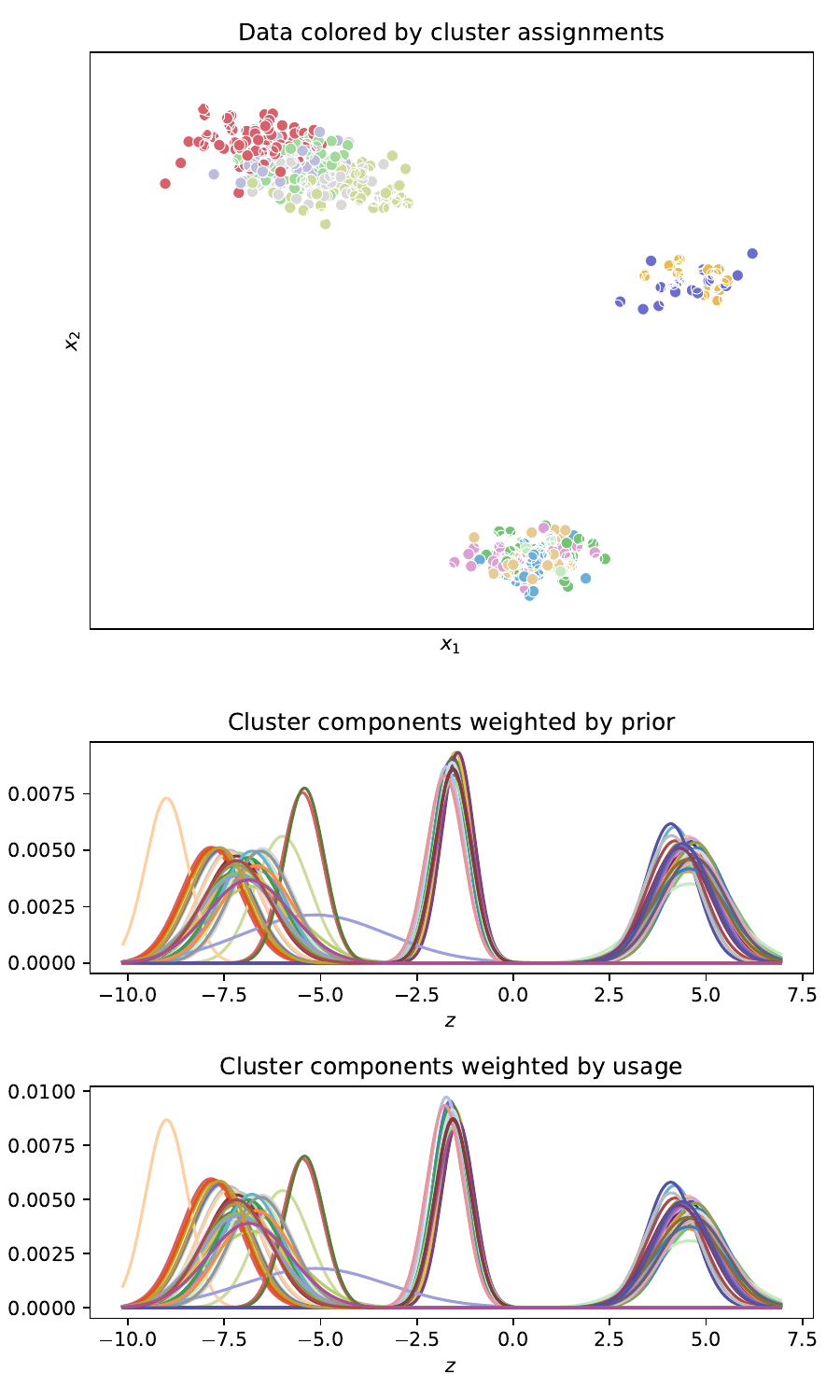}
    \caption{VampPrior-Unleashed}
    \label{fig:toy:unleashed}
  \end{subfigure}
  \hfill
  \begin{subfigure}[t]{0.3\textwidth}
    \centering
    \includegraphics[width=\textwidth]{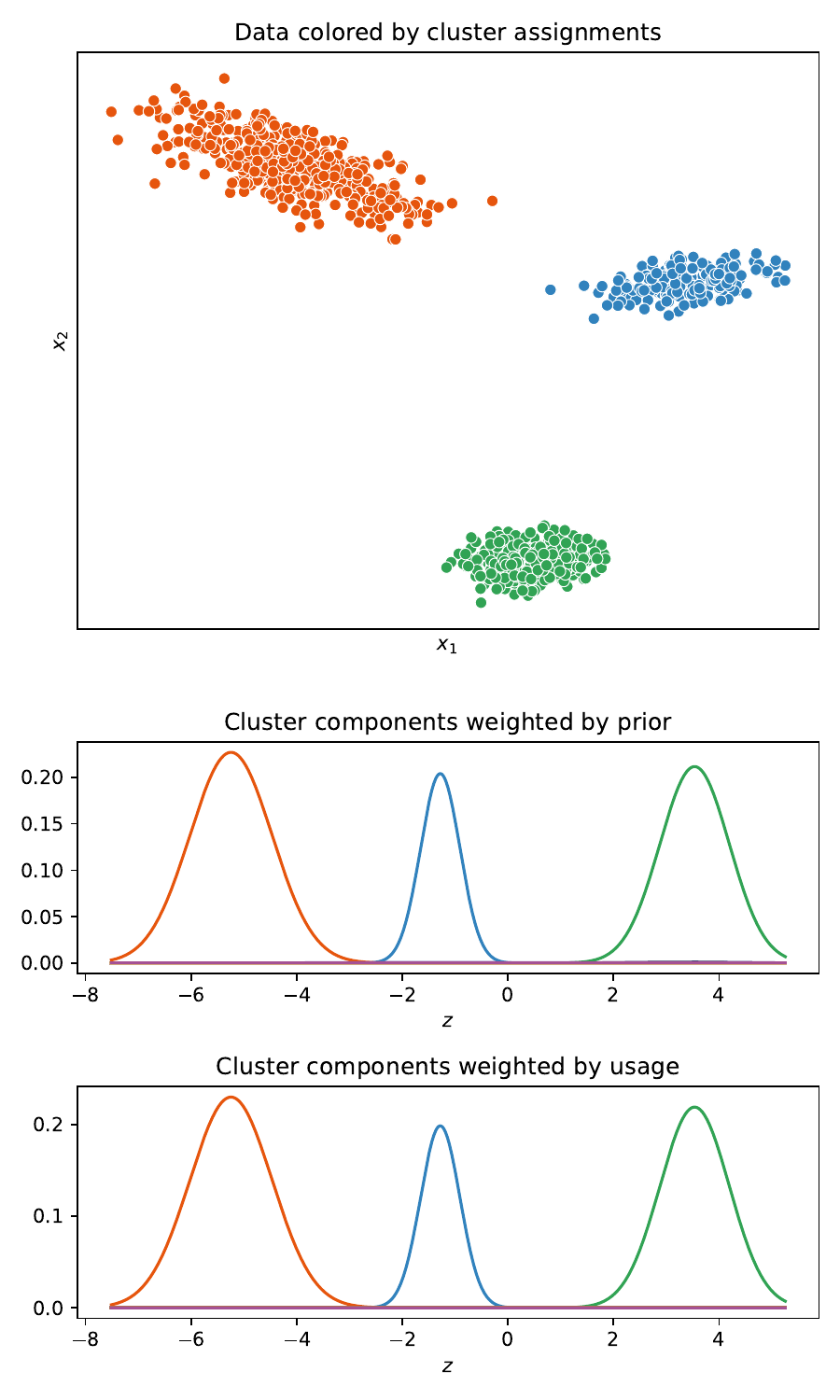}
    \caption{DP VampPrior-Unleashed}
    \label{fig:toy:dp-unleashed}
  \end{subfigure}
  \caption{Modeling a 2D anisotropic GMM (three well-separated components) with a 1D latent and $K=40$ available components. Top: data colored by cluster assignment. Middle and bottom: the prior's component densities weighted by the mixture proportions $\pi$ and by realized test-set usage, respectively. Only our DP VampPrior-Unleashed (\subref{fig:toy:dp-unleashed}) recovers the three true clusters.}
  \label{fig:toy}
\end{figure*}

\begin{figure*}
  \centering
  \includegraphics[width=\textwidth]{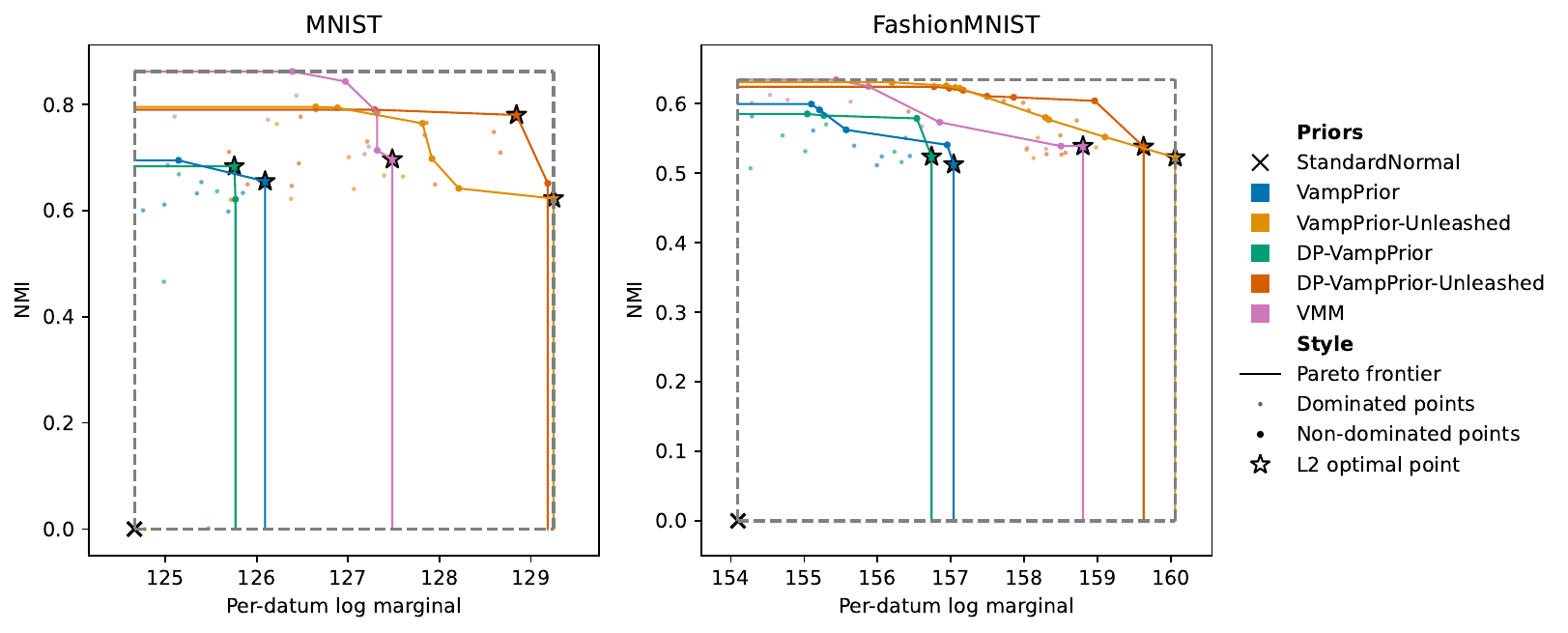}
  \caption{Evidence--NMI Pareto frontiers. Each point is one hyperparameter configuration averaged over $20$ seeds; large points are the non-dominated configurations and the line connecting them interpolates the Pareto frontier. The dashed box spans the standard normal VAE (origin, $\times$) to the best per-datum log marginal and NMI attained by any prior. Configurations whose mean log marginal or NMI falls outside the grey box, i.e., worse than the standard normal VAE on either axis, are excluded before the frontier is computed. Stars mark each prior's $P^{\ast}$, the frontier point closest to the normalized ideal (top-right corner of grey box after normalization to unit box). Both our VampPrior-Unleashed and DP-VampPrior-Unleashed offer sizable frontier advancements.}
  \label{fig:pareto_figure}
\end{figure*}

\begin{table*}
  \caption{Pareto summary statistics with $[2.5,97.5]$ bootstrap intervals over $10{,}000$ samples. Boldface marks the best prior per column together with every prior statistically indistinguishable from it ($0.025<P(\Delta>0)<0.975$ in \cref{tab:pareto_diffs}, \cref{supp:pareto-diffs}).}
  \label{tab:pareto_means}
  \begin{center}
    \adjustbox{width=\textwidth}{\begin{tabular}{llllll}
\toprule
Dataset & Prior & Hypervolume & Min $||P^{\ast} - (1,1)||_2$ & Max Log marginal & Max NMI \\
\midrule
\multirow[t]{5}{*}{MNIST} & VampPrior & 0.242 [0.115, 0.423] & 0.729 [0.517, 0.887] & 126.092 [125.740, 127.199] & 0.695 [0.654, 0.701] \\
 & VampPrior-Unleashed & \textbf{0.856 [0.677, 0.883]} & \textbf{0.278 [0.178, 0.347]} & \textbf{129.249 [128.494, 130.120]} & 0.796 [0.788, 0.810] \\
 & DP-VampPrior & 0.191 [0.066, 0.380] & 0.789 [0.569, 0.939] & 125.768 [125.501, 126.876] & 0.683 [0.664, 0.693] \\
 & DP-VampPrior-Unleashed & \textbf{0.888 [0.741, 0.917]} & \textbf{0.130 [0.091, 0.260]} & \textbf{129.185 [128.752, 130.128]} & 0.790 [0.786, 0.802] \\
 & VMM & 0.599 [0.482, 0.820] & 0.430 [0.203, 0.523] & 127.484 [127.320, 128.762] & \textbf{0.862 [0.856, 0.867]} \\
\cline{1-6}
\multirow[t]{5}{*}{FashionMNIST} & VampPrior & 0.439 [0.359, 0.556] & 0.542 [0.409, 0.623] & 157.038 [156.678, 157.780] & 0.599 [0.597, 0.602] \\
 & VampPrior-Unleashed & \textbf{0.933 [0.846, 0.952]} & \textbf{0.176 [0.130, 0.201]} & \textbf{160.055 [159.461, 160.691]} & \textbf{0.631 [0.625, 0.644]} \\
 & DP-VampPrior & 0.403 [0.333, 0.522] & 0.584 [0.448, 0.652] & 156.737 [156.483, 157.443] & 0.585 [0.581, 0.591] \\
 & DP-VampPrior-Unleashed & \textbf{0.889 [0.765, 0.963]} & \textbf{0.168 [0.087, 0.246]} & \textbf{159.625 [159.020, 160.520]} & \textbf{0.624 [0.620, 0.633]} \\
 & VMM & 0.723 [0.628, 0.840] & 0.259 [0.172, 0.351] & 158.800 [158.412, 159.463] & \textbf{0.635 [0.630, 0.639]} \\
\cline{1-6}
\bottomrule
\end{tabular}
}
  \end{center}
\end{table*}

Improved evidence does not necessarily make a prior more useful for downstream tasks that motivate DLVMs in single-cell analysis, in particular clustering.
In \cref{sec:experiments:unleashing}, the number of active components for both the \textbf{VampPrior} and the \textbf{VampPrior-Unleashed} is always in $\{K-1,K\}$.
Accordingly, NMI consistently peaks near $K=10$--$20$ and then decays monotonically.
While decoupling shifts the whole NMI curve upward, it leaves its shape intact, because the pathology lies with the fixed uniform mixture weights and not the component variances.
\Cref{fig:toy} makes this failure mode concrete on a 2D anisotropic GMM with three well-separated components modeled by a 1D latent with $K=40$ available components.
Both the \textbf{VampPrior} (\cref{fig:toy:original}) and the \textbf{VampPrior-Unleashed} (\cref{fig:toy:unleashed}) spend all $40$ components, fragmenting each true cluster into a haze of arbitrary sub-assignments; the usage-weighted densities show no component going unused.
The \textbf{DP VampPrior-Unleashed} (\cref{fig:toy:dp-unleashed}) prunes the same $40$ available components down to exactly three, one per true cluster, and recovers the generative partition.

\Cref{fig:sweep_alpha:fashion_mnist,fig:sweep_alpha:mnist} (\cref{supp:sweep_alpha}) sweep the concentration $\alpha$ with the truncation held at $K=100$ and isolate the effect of covariance decoupling in the DP setting.
As with the $K$ sweeps (\cref{fig:sweep_k:fashion_mnist:data,fig:sweep_k:fashion_mnist,fig:sweep_k:mnist}), our decoupled \textbf{DP VampPrior-Unleashed} consistently outperforms its non-decoupled counterpart on the same panel of metrics across a wide range of $\alpha$ values on all datasets and under all $u$ initializations.
The typical per-datum log marginal gain is $+2$ to $+4$ nats indicating a decisive improvement in model evidence.
Furthermore, $\alpha$ is the parsimony dial the DP promises; e.g., in \cref{fig:sweep_alpha:mnist:random} our \textbf{DP VampPrior-Unleashed}  on average activates $1$ component at $\alpha=0.1$, $6.6$ components at $\alpha=0.5$, $15.8$ at $\alpha=1$, $40.6$ at $\alpha=2$, and $99.2$ at $\alpha=10$.

\subsection{Trading Evidence for Structure}
\label{sec:experiments:pareto}

\Cref{sec:experiments:unleashing,sec:experiments:clustering} establish that decoupling helps at fixed hyperparameter values, but each sections' priors have their own hyperparameter axis ($K$ and $\alpha$).
To compare priors across these different axes, we construct an evidence--clustering plane, where the set of configurations tested for each prior produce a set of points.
The VMM, where we mimic \textcite{stirnVampPriorMixtureModel2025}'s parameterization via batch size, can also be analyzed in this plane.
\Cref{fig:pareto_figure} plots every configuration's (per-datum log marginal, NMI) point and traces the Pareto frontier for each prior.
\Cref{tab:pareto_means} quantitatively summarizes those frontiers.
We normalize the plane to a unit square (grey box, \cref{fig:pareto_figure}) where the bottom-left corner is the origin.
Hypervolume is the area covered by the staircase integral of non-dominated points---a conservative estimate for the true Pareto frontier as estimated by the solid lines in \cref{fig:pareto_figure}.
$P^{\ast}$ is the frontier point closest to the normalized ideal $(1,1)$---top-right corner of the grey box.
The minimum distance from the normalized ideal $(1,1)$ indicates how close a prior's Pareto frontier can get to this ideal.

Both decoupled variants dominate their coupled counterparts on hypervolume and L$_2$ distance to the ideal (\cref{tab:pareto_means,tab:pareto_diffs}).
Decoupling more than triples hypervolume on MNIST: from $0.242$ to $0.856$ for the uniform mixture and from $0.191$ to $0.888$ for the DP mixture.
The coupled DP is the weakest prior we tested on both datasets and is statistically indistinguishable from the original \textbf{VampPrior}, confirming our covariance decoupling is what makes the DP viable.

Comparison to the \textbf{VMM} is more nuanced.
Our \textbf{DP-VampPrior-Unleashed} attains a larger hypervolume and a closer best-compromise point on both datasets.
Its maximum per-datum evidence exceeds the \textbf{VMM}'s by $1.7$ nats on MNIST and $0.8$ on FashionMNIST, delivering decisive Bayes factors.
Crucially, at the best-compromise point $P^{\ast}$, the two priors' NMIs are statistically indistinguishable on both datasets, while the \textbf{DP-VampPrior-Unleashed}'s $P^{\ast}$ log marginal is significantly higher on MNIST (\cref{tab:pareto_diffs}).
In other words, at an operating point that one might pragmatically select, our prior matches the \textbf{VMM}'s clustering while extracting materially more evidence from the same architecture.
The \textbf{VMM} nonetheless retains the highest achievable NMI, which \textcite{stirnVampPriorMixtureModel2025} explicitly designed it for.
Yet, on FashionMNIST, the margin is negligible relative to our \textbf{VampPrior-Unleashed} and \textbf{DP-VampPrior-Unleashed}.
However, on MNIST the \textbf{VMM}'s $0.862$ NMI is a genuine win.
We regard this as the honest form of our contribution: our proposals are not uniformly best at clustering but offer a strictly better Pareto frontier.

Comparing our decoupled variants to each other, the DP mixture is statistically indistinguishable from the uniform mixture on every frontier statistic and both datasets ($0.064\leq P(\Delta>0)\leq0.970$).
This result is expected: the uniform mixture reaches its frontier by tuning $K$, and its NMI collapses as $K$ grows too large, requiring the number of clusters to be known a priori.
In contrast, the DP reaches the same frontier from a single generous truncation of $K=100$, pruning to as few as $16$ active components on its own with reasonable $\alpha$ values.
After all, matching a tuned model without tuning is the goal of Bayesian nonparametric modeling.

\subsection{scRNA-seq}
\label{sec:experiments:scRNA}

\begin{table*}
  \caption{Summary statistics for scRNA-seq experiments. Entries are means $\pm1$ standard deviation over $3$ seeds. Boldface marks, per column, the best mean and every prior tied with it at the reported precision; this is a descriptive convention, not a significance claim.}
  \label{tab:scRNA}
  \begin{center}
    \adjustbox{width=\textwidth}{\begin{tabular}{lll|ll|lllll|llll|lll}
\toprule
 &  & Encode & \multicolumn{2}{|c}{Our metrics} & \multicolumn{5}{|c}{Bio conservation} & \multicolumn{4}{|c}{Batch correction} & \multicolumn{3}{|c}{Aggregate} \\
Dataset & Prior & covariates & Log Marginal & NMI & Isolated labels & KMeans ARI & KMeans NMI & Silhouette label & cLISI & BRAS & Graph connectivity & KBET & iLISI & Batch correction & Bio conservation & Total \\
\midrule
\multirow[t]{4}{*}{DKD} & \multirow[t]{2}{*}{StandardNormal} & False & -763.596 $\pm$ 0.61 & 0.000 $\pm$ 0.00 & 0.617 $\pm$ 0.00 & 0.497 $\pm$ 0.05 & 0.717 $\pm$ 0.03 & 0.588 $\pm$ 0.00 & \textbf{1.000 $\pm$ 0.00} & 0.845 $\pm$ 0.00 & 0.921 $\pm$ 0.00 & 0.299 $\pm$ 0.00 & 0.277 $\pm$ 0.00 & 0.585 $\pm$ 0.00 & 0.684 $\pm$ 0.02 & 0.645 $\pm$ 0.01 \\
 &  & True & -763.285 $\pm$ 0.60 & 0.000 $\pm$ 0.00 & 0.620 $\pm$ 0.00 & 0.526 $\pm$ 0.02 & 0.726 $\pm$ 0.02 & 0.587 $\pm$ 0.00 & \textbf{1.000 $\pm$ 0.00} & \textbf{0.865 $\pm$ 0.00} & 0.918 $\pm$ 0.00 & 0.306 $\pm$ 0.01 & 0.270 $\pm$ 0.00 & 0.590 $\pm$ 0.00 & 0.692 $\pm$ 0.01 & 0.651 $\pm$ 0.00 \\
 & VMM & True & -764.153 $\pm$ 0.27 & 0.457 $\pm$ 0.01 & 0.670 $\pm$ 0.01 & 0.623 $\pm$ 0.09 & 0.757 $\pm$ 0.04 & 0.655 $\pm$ 0.00 & \textbf{1.000 $\pm$ 0.00} & 0.755 $\pm$ 0.02 & 0.941 $\pm$ 0.00 & 0.380 $\pm$ 0.01 & 0.320 $\pm$ 0.00 & 0.599 $\pm$ 0.01 & 0.741 $\pm$ 0.02 & 0.684 $\pm$ 0.01 \\
 & DP-VampPrior-Unleashed & True & \textbf{-762.546 $\pm$ 0.57} & \textbf{0.762 $\pm$ 0.01} & \textbf{0.683 $\pm$ 0.00} & \textbf{0.666 $\pm$ 0.09} & \textbf{0.779 $\pm$ 0.01} & \textbf{0.694 $\pm$ 0.00} & \textbf{1.000 $\pm$ 0.00} & 0.764 $\pm$ 0.01 & \textbf{0.942 $\pm$ 0.00} & \textbf{0.406 $\pm$ 0.01} & \textbf{0.332 $\pm$ 0.00} & \textbf{0.611 $\pm$ 0.00} & \textbf{0.764 $\pm$ 0.02} & \textbf{0.703 $\pm$ 0.01} \\
\midrule
\multirow[t]{3}{*}{Immune Cell Atlas} & \multirow[t]{2}{*}{StandardNormal} & False & -548.564 $\pm$ 1.59 & 0.000 $\pm$ 0.00 & 0.591 $\pm$ 0.01 & 0.475 $\pm$ 0.03 & 0.689 $\pm$ 0.01 & 0.543 $\pm$ 0.01 & 0.999 $\pm$ 0.00 & 0.807 $\pm$ 0.00 & 0.924 $\pm$ 0.00 & 0.183 $\pm$ 0.00 & 0.094 $\pm$ 0.00 & 0.502 $\pm$ 0.00 & 0.659 $\pm$ 0.01 & 0.596 $\pm$ 0.00 \\
 &  & True & -548.226 $\pm$ 1.38 & 0.000 $\pm$ 0.00 & 0.614 $\pm$ 0.00 & 0.488 $\pm$ 0.02 & 0.684 $\pm$ 0.00 & 0.540 $\pm$ 0.00 & 0.999 $\pm$ 0.00 & \textbf{0.825 $\pm$ 0.00} & 0.922 $\pm$ 0.00 & 0.179 $\pm$ 0.00 & 0.098 $\pm$ 0.00 & 0.506 $\pm$ 0.00 & 0.665 $\pm$ 0.00 & 0.601 $\pm$ 0.00 \\
 & DP-VampPrior-Unleashed & True & \textbf{-545.786 $\pm$ 1.41} & \textbf{0.787 $\pm$ 0.01} & \textbf{0.659 $\pm$ 0.03} & \textbf{0.634 $\pm$ 0.03} & \textbf{0.764 $\pm$ 0.01} & \textbf{0.595 $\pm$ 0.00} & \textbf{1.000 $\pm$ 0.00} & 0.787 $\pm$ 0.00 & \textbf{0.933 $\pm$ 0.00} & \textbf{0.241 $\pm$ 0.00} & \textbf{0.129 $\pm$ 0.00} & \textbf{0.522 $\pm$ 0.00} & \textbf{0.730 $\pm$ 0.02} & \textbf{0.647 $\pm$ 0.01} \\
\midrule
\multirow[t]{4}{*}{GTEx} & \multirow[t]{2}{*}{StandardNormal} & False & -383.745 $\pm$ 0.79 & 0.000 $\pm$ 0.00 & 0.668 $\pm$ 0.01 & 0.356 $\pm$ 0.01 & 0.677 $\pm$ 0.00 & 0.531 $\pm$ 0.00 & \textbf{1.000 $\pm$ 0.00} & \textbf{0.745 $\pm$ 0.01} & 0.879 $\pm$ 0.01 & 0.280 $\pm$ 0.01 & 0.046 $\pm$ 0.00 & \textbf{0.488 $\pm$ 0.00} & 0.646 $\pm$ 0.00 & 0.583 $\pm$ 0.00 \\
 &  & True & -383.449 $\pm$ 0.81 & 0.000 $\pm$ 0.00 & \textbf{0.680 $\pm$ 0.01} & 0.330 $\pm$ 0.01 & 0.654 $\pm$ 0.01 & 0.525 $\pm$ 0.00 & \textbf{1.000 $\pm$ 0.00} & 0.718 $\pm$ 0.01 & \textbf{0.885 $\pm$ 0.01} & 0.220 $\pm$ 0.02 & 0.021 $\pm$ 0.00 & 0.461 $\pm$ 0.01 & 0.637 $\pm$ 0.01 & 0.567 $\pm$ 0.01 \\
 & VMM & True & -383.547 $\pm$ 0.00 & 0.000 $\pm$ 0.00 & 0.675 $\pm$ nan & 0.315 $\pm$ nan & 0.637 $\pm$ nan & 0.512 $\pm$ nan & 0.999 $\pm$ nan & 0.691 $\pm$ nan & 0.855 $\pm$ nan & 0.260 $\pm$ nan & 0.047 $\pm$ nan & 0.463 $\pm$ nan & 0.628 $\pm$ nan & 0.562 $\pm$ nan \\
 & DP-VampPrior-Unleashed & True & \textbf{-381.235 $\pm$ 0.75} & \textbf{0.784 $\pm$ 0.01} & 0.673 $\pm$ 0.03 & \textbf{0.463 $\pm$ 0.05} & \textbf{0.730 $\pm$ 0.00} & \textbf{0.556 $\pm$ 0.01} & \textbf{1.000 $\pm$ 0.00} & 0.707 $\pm$ 0.01 & 0.868 $\pm$ 0.02 & \textbf{0.297 $\pm$ 0.01} & \textbf{0.067 $\pm$ 0.00} & 0.485 $\pm$ 0.00 & \textbf{0.684 $\pm$ 0.00} & \textbf{0.604 $\pm$ 0.00} \\
\midrule
\multirow[t]{4}{*}{Mouse Pancreas Atlas} & \multirow[t]{2}{*}{StandardNormal} & False & -688.704 $\pm$ 0.89 & 0.000 $\pm$ 0.00 & 0.716 $\pm$ 0.01 & 0.446 $\pm$ 0.03 & 0.646 $\pm$ 0.02 & 0.526 $\pm$ 0.00 & \textbf{1.000 $\pm$ 0.00} & 0.822 $\pm$ 0.00 & \textbf{0.924 $\pm$ 0.00} & 0.250 $\pm$ 0.01 & 0.048 $\pm$ 0.00 & 0.511 $\pm$ 0.00 & 0.667 $\pm$ 0.01 & 0.605 $\pm$ 0.01 \\
 &  & True & \textbf{-687.781 $\pm$ 0.86} & 0.000 $\pm$ 0.00 & 0.714 $\pm$ 0.01 & 0.440 $\pm$ 0.04 & 0.640 $\pm$ 0.02 & 0.530 $\pm$ 0.00 & \textbf{1.000 $\pm$ 0.00} & \textbf{0.858 $\pm$ 0.00} & 0.915 $\pm$ 0.01 & 0.261 $\pm$ 0.01 & 0.057 $\pm$ 0.00 & 0.523 $\pm$ 0.00 & 0.665 $\pm$ 0.01 & 0.608 $\pm$ 0.01 \\
 & VMM & True & -690.579 $\pm$ 0.00 & 0.599 $\pm$ 0.00 & \textbf{0.785 $\pm$ nan} & 0.614 $\pm$ nan & 0.732 $\pm$ nan & 0.600 $\pm$ nan & \textbf{1.000 $\pm$ nan} & 0.823 $\pm$ nan & 0.896 $\pm$ nan & 0.344 $\pm$ nan & 0.087 $\pm$ nan & 0.537 $\pm$ nan & 0.746 $\pm$ nan & 0.663 $\pm$ nan \\
 & DP-VampPrior-Unleashed & True & -687.858 $\pm$ 1.05 & \textbf{0.788 $\pm$ 0.01} & 0.752 $\pm$ 0.01 & \textbf{0.671 $\pm$ 0.11} & \textbf{0.758 $\pm$ 0.03} & \textbf{0.632 $\pm$ 0.01} & \textbf{1.000 $\pm$ 0.00} & 0.818 $\pm$ 0.00 & 0.889 $\pm$ 0.00 & \textbf{0.378 $\pm$ 0.01} & \textbf{0.089 $\pm$ 0.00} & \textbf{0.543 $\pm$ 0.00} & \textbf{0.763 $\pm$ 0.03} & \textbf{0.675 $\pm$ 0.02} \\
\midrule
\multirow[t]{3}{*}{HypoMap} & \multirow[t]{2}{*}{StandardNormal} & False & -684.303 $\pm$ 1.65 & 0.000 $\pm$ 0.00 & \textbf{0.397 $\pm$ 0.01} & 0.246 $\pm$ 0.01 & 0.592 $\pm$ 0.02 & 0.533 $\pm$ 0.00 & \textbf{1.000 $\pm$ 0.00} & 0.827 $\pm$ 0.01 & \textbf{0.872 $\pm$ 0.00} & 0.868 $\pm$ 0.00 & \textbf{0.000 $\pm$ 0.00} & 0.642 $\pm$ 0.00 & 0.554 $\pm$ 0.01 & 0.589 $\pm$ 0.00 \\
 &  & True & -684.165 $\pm$ 1.55 & 0.000 $\pm$ 0.00 & 0.392 $\pm$ 0.01 & 0.226 $\pm$ 0.04 & 0.559 $\pm$ 0.06 & 0.530 $\pm$ 0.00 & \textbf{1.000 $\pm$ 0.00} & \textbf{0.831 $\pm$ 0.00} & 0.850 $\pm$ 0.00 & \textbf{0.904 $\pm$ 0.00} & \textbf{0.000 $\pm$ 0.00} & \textbf{0.646 $\pm$ 0.00} & 0.542 $\pm$ 0.02 & 0.583 $\pm$ 0.01 \\
 & DP-VampPrior-Unleashed & True & \textbf{-681.795 $\pm$ 1.55} & \textbf{0.669 $\pm$ 0.02} & 0.309 $\pm$ 0.01 & \textbf{0.356 $\pm$ 0.11} & \textbf{0.663 $\pm$ 0.04} & \textbf{0.638 $\pm$ 0.01} & \textbf{1.000 $\pm$ 0.00} & 0.777 $\pm$ 0.01 & 0.810 $\pm$ 0.02 & 0.878 $\pm$ 0.01 & \textbf{0.000 $\pm$ 0.00} & 0.616 $\pm$ 0.00 & \textbf{0.593 $\pm$ 0.03} & \textbf{0.602 $\pm$ 0.02} \\
\midrule
\multirow[t]{3}{*}{Tabula Sapiens} & \multirow[t]{2}{*}{StandardNormal} & False & -778.511 $\pm$ 1.85 & 0.000 $\pm$ 0.00 & \textbf{0.581 $\pm$ 0.00} & 0.265 $\pm$ 0.01 & 0.673 $\pm$ 0.00 & 0.496 $\pm$ 0.00 & \textbf{1.000 $\pm$ 0.00} & \textbf{0.679 $\pm$ 0.01} & 0.735 $\pm$ 0.00 & 0.345 $\pm$ 0.01 & 0.008 $\pm$ 0.00 & 0.442 $\pm$ 0.00 & 0.603 $\pm$ 0.00 & 0.538 $\pm$ 0.00 \\
 &  & True & -778.099 $\pm$ 1.52 & 0.000 $\pm$ 0.00 & 0.578 $\pm$ 0.00 & 0.258 $\pm$ 0.01 & 0.666 $\pm$ 0.00 & \textbf{0.498 $\pm$ 0.00} & \textbf{1.000 $\pm$ 0.00} & 0.677 $\pm$ 0.01 & \textbf{0.759 $\pm$ 0.00} & 0.340 $\pm$ 0.01 & 0.004 $\pm$ 0.00 & \textbf{0.445 $\pm$ 0.01} & 0.600 $\pm$ 0.00 & 0.538 $\pm$ 0.00 \\
 & DP-VampPrior-Unleashed & True & \textbf{-772.523 $\pm$ 1.58} & \textbf{0.714 $\pm$ 0.00} & 0.578 $\pm$ 0.00 & \textbf{0.283 $\pm$ 0.01} & \textbf{0.692 $\pm$ 0.00} & \textbf{0.498 $\pm$ 0.00} & \textbf{1.000 $\pm$ 0.00} & 0.640 $\pm$ 0.00 & 0.730 $\pm$ 0.01 & \textbf{0.365 $\pm$ 0.01} & \textbf{0.009 $\pm$ 0.00} & 0.436 $\pm$ 0.00 & \textbf{0.610 $\pm$ 0.00} & \textbf{0.540 $\pm$ 0.00} \\
\bottomrule
\end{tabular}
}
  \end{center}
\end{table*}

The canonical DLVM for scRNA-seq is scVI \parencite{lopezDeepGenerativeModeling2018},
a VAE that conditions on a batch label $s_i$ (e.g., donor or patient) and uses a zero-inflated negative binomial (ZINB) likelihood suitable for sparse, over-dispersed scRNA-seq counts.
For the $i$-th cell, scVI observes read counts $x_i\in\mathbb{N}^G$, a library size $l_i\triangleq\sum_j x_{ij}$, and batch label $s_i$ and posits,
\begin{align*}
  p(Z)                           & \triangleq \N(0,I)                                  \\
  p_{\theta}(X \mid Z=z,L=l,S=s) & \triangleq \text{ZINB}\big(X;f_{\theta}(z,s),l\big) \\
  q_{\phi}(Z \mid X=x,S=s)       & \triangleq \N(Z;g_{\phi}(x,s)).
\end{align*}
We replace scVI's standard normal prior with our \textbf{DP VampPrior-Unleashed} and leave the encoder, decoder, and likelihood untouched, so that any change in performance is attributable to the prior alone.
We favor our DP variant for its demonstrated ability to preserve latent structure across a wide range of concentrations (\cref{sec:experiments:pareto}) without committing to the number of available components a priori.

For our scRNA-seq evaluation, we follow \href{https://openproblems.bio/}{OpenProblems.bio}.
We use the same six publicly available atlases and report scIB metrics \parencite{lueckenBenchmarkingAtlaslevelData2022} alongside our measured log marginal and NMI in \cref{tab:scRNA}.
See \cref{supp:experimental-setup:metrics:scib} for a summary of the scIB metrics.
Documentation for scVI notes little difference in providing the batch label to the encoder or not (encode covariates column, \cref{tab:scRNA})---the decoder always sees the batch label.
Recognizing the true posterior has this dependence, we test scVI's standard-normal baseline in both settings and our \textbf{DP VampPrior-Unleashed} with covariate encoding enabled.

Most notably, our method improves scIB's aggregate total score on \emph{every} atlas by increasing biological conservation as measured by the underlying metrics and their aggregate.
KMeans adjusted Rand Index (ARI)/NMI and silhouette-label are the big drivers of this improvement under our prior.
These metrics, particularly benefit from enhanced biological structure captured in the latent space, which our \textbf{DP VampPrior-Unleashed} is designed to preserve.
Crucially, this structure does not come at the expense of model evidence.
Our prior attains a higher per-datum log marginal on five of six atlases---by up to $+5.6$ nats on Tabula Sapiens, and by $+2$ to $+2.5$ nats on the Immune Cell Atlas, GTEx, and HypoMap---and appears statistically indistinguishable on the Mouse Pancreas Atlas.
Exponentiated over tens to hundreds of thousands of cells, these differences are decisive Bayes factors, mirroring the zero-tradeoff behavior in \cref{sec:experiments:unleashing}.
The VMM only completed all three seeds without hitting NaNs on DKD.
Its other two appearances have $\pm$ NaN as the standard error, indicating only one seed completed successfully.

The one qualified exception may be batch correction, where the aggregate score is mixed and the pattern is metric-specific.
kBET and iLISI, which reward batch mixing \emph{within} a cell type, improve with our prior or tie on nearly every atlas, whereas BRAS and graph connectivity favor the standard normal throughout.
A diffuse, structureless embedding trivially maximizes BRAS---since within-cell-type distances to same-batch and other-batch cells become indistinguishable---and similarly satisfies graph connectivity by collapsing cells into a single connected component regardless of biological structure.
The authors of BRAS \parencite{rautenstrauchShortcomingsSilhouetteSinglecell2026} dispute the reliability of silhouette-based metrics for single-cell batch correction more broadly and note that BRAS remains sensitive to differential cell type composition between donors and should not be interpreted in isolation.
\Cref{tab:scrna_datasets} (\cref{supp:scrna}) computes Cramér's $V$ for each dataset in \cref{tab:scRNA} and finds a high degree of association between cells and donors for two of the three datasets where our \textbf{DP VampPrior-Unleashed} did not achieve the best batch correction score.
In other words, the standard normal prior appears to earn its winning batch-correction scores by discarding biological structure our prior is designed to retain.


\section{Conclusion}

Motivated by DLVMs widespread adoption in modern biological data analyses and the importance prior selection has, we seek a prior that provides optimal balance between model evidence and biological structure identification.
Starting with the VampPrior, which approximates the optimal prior, we identify two key limitations that prohibit it from meeting these desiderata.
First, we observe that the VampPrior forces a single amortized covariance to serve both a low-entropy posterior and a high-entropy prior, and that decoupling the component covariances from the encoder (our \emph{VampPrior-Unleashed}) substantially raises model evidence by simultaneously improving rate \emph{and} distortion.
We then replace the uniform mixture of prior components with a Dirichlet process, yielding our \emph{DP-VampPrior-Unleashed}, which prunes redundant components and recovers parsimonious, well-separated clusters and offers the best evidence--structure Pareto frontier via a single concentration parameter, thereby avoiding fixing the number of clusters a priori.
Our \emph{DP-VampPrior-Unleashed} when used within scVI substantially improves both model evidence and biological structure recovery across diverse single-cell atlases.

Our contributions apply to any continuous latent-space DLVM, and several extensions are natural.
Expanding the \textbf{DP VampPrior-Unleashed} to a hierarchical Dirichlet process would share components across groups, providing patient- or sample-specific mixture proportions for multi-sample atlases and topic-model-style analyses.
More broadly, because the prior is the only component we change, we anticipate the same evidence and structure gains for the many DLVMs---across ATAC-seq, methylation, multi-omics, and spatial transcriptomics---that often default to a standard normal or a fixed-size GMM prior.

\clearpage


\printbibliography

\clearpage
\appendix
\thispagestyle{empty}

\onecolumn
\aistatstitle{VampPrior Unleashed: Supplementary Materials}

\section{Prior Lineage}
\label{supp:prior-lineage}

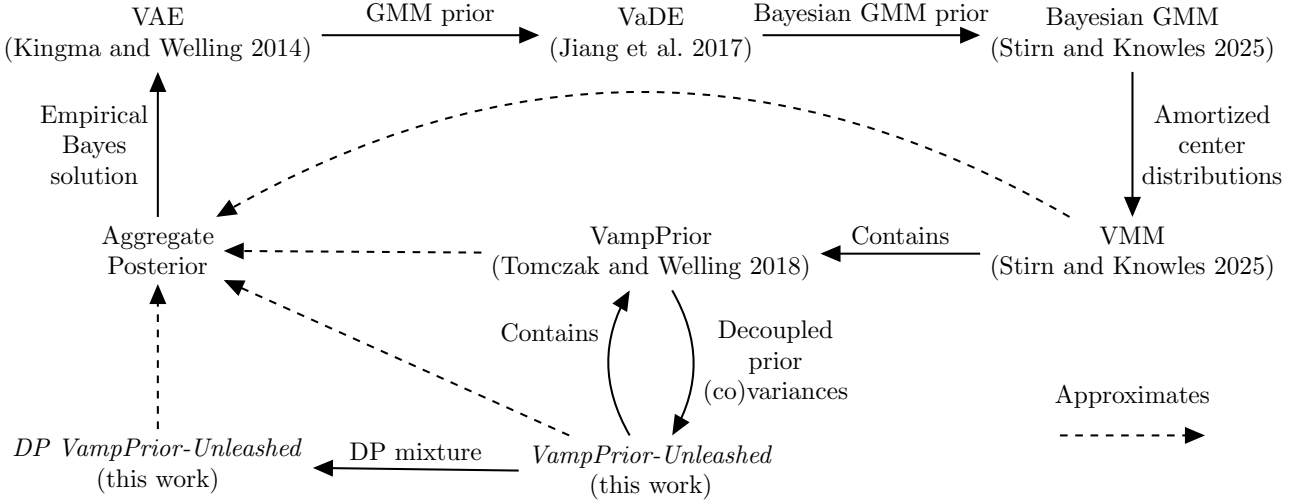
\begin{figure}[H]
  \centering
  \begin{tikzpicture}[->,thick, every text node part/.style={align=center}]

    \node (VAE) {VAE \\ \parencite{kingmaAutoEncodingVariationalBayes2014}};
    \node[right=3cm of VAE] (VaDE) {VaDE \\ \parencite{jiangVariationalDeepEmbedding2017}};
    \node[right=3cm of VaDE] (Bayesian-GMM) {Bayesian GMM \\ \parencite{stirnVampPriorMixtureModel2025}};
    \node[below=2cm of VAE] (AggPost) {Aggregate \\ Posterior};
    \node[below=2cm of VaDE] (VampPrior) {VampPrior \\ \parencite{tomczakVAEVampPrior2018}};
    \node[below=2cm of Bayesian-GMM] (VMM) {VMM \\ \parencite{stirnVampPriorMixtureModel2025}};
    \node[below=2cm of VampPrior] (VampPrior-Unleashed) {\emph{VampPrior-Unleashed} \\ (this work)};
    \node[below=2cm of AggPost] (DP-VampPrior-Unleashed) {\emph{DP VampPrior-Unleashed} \\ (this work)};

    \path (AggPost) edge node[left] {Empirical \\ Bayes \\ solution} (VAE);
    \path (VAE) edge node[above] {GMM prior} (VaDE);
    \path (VaDE) edge node[above] {Bayesian GMM prior} (Bayesian-GMM);
    \path (Bayesian-GMM) edge node[right] {Amortized \\ center \\ distributions} (VMM);
    \path (VampPrior) edge[dashed] (AggPost);
    \path (VMM) edge node[above] {Contains} (VampPrior);
    \path (VMM) edge[dashed, bend right] (AggPost);
    \path (VampPrior) edge[bend left] node[right] {Decoupled \\ prior \\ (co)variances} (VampPrior-Unleashed);
    \path (VampPrior-Unleashed) edge[bend left] node[left, yshift=12pt] {Contains} (VampPrior);
    \path (VampPrior-Unleashed) edge[dashed] (AggPost);
    \path (VampPrior-Unleashed) edge node[above] {DP mixture} (DP-VampPrior-Unleashed);
    \path (DP-VampPrior-Unleashed) edge[dashed] (AggPost);

    \draw[dashed] ([shift={(-1cm,-2cm)}]VMM.south) -- ([shift={(1cm,-2cm)}]VMM.south) node[midway, above=0.25cm] {Approximates};

\end{tikzpicture}
  \caption{Related work: a lineage of Variational Autoencoder (VAE) priors.}
  \label{fig:vae_priors}
\end{figure}

\Cref{fig:vae_priors}, which \cref{sec:methods:related-work} references, provides an overview of the evolution of VAE priors.
Two branches emerge from a VAE with a standard normal prior: the top row replaces $p(Z)$ with increasingly flexible clustering priors, whereas the VampPrior approximates the aggregate posterior via \cref{eq:vampprior}.
The VMM joins the two branches.
Our \emph{VampPrior-Unleashed} and \emph{DP-VampPrior-Unleashed} extend the VampPrior branch by decoupling prior variance from the encoder and, in the case of the latter, by adding a DP mixture.
Arrows labeled ``Contains'' denote that the target is recovered as a special case.

\section{Experimental Setup}
\label{supp:experimental-setup}

In \cref{sec:experiments:unleashing,sec:experiments:clustering,sec:experiments:pareto}, every model shares an identical encoder, decoder, latent dimension, likelihood, and optimization schedule, so that all differences are attributable to the prior alone.
Furthermore, a seed ensures identical initialization and data shuffling across models, allowing for paired comparisons between models.
For both FashionMNIST and MNIST, we use an architecture consistent with the VAE clustering literature \parencite{jiangVariationalDeepEmbedding2017,stirnVampPriorMixtureModel2025}.
The encoder is a three-layer MLP ($500 \to 500 \to 2000$, with ELU activations) mapping onto a diagonal Gaussian.
The decoder is the reversed three-layer MLP ($2000 \to 500 \to 500 \to \dim_X$, with ELU activations) mapping onto an isotropic Gaussian $p_{\theta}(X \mid Z=z)$ with a learned global variance.
FashionMNIST and MNIST use 10 and 15 latent dimensions, respectively.

In \cref{sec:experiments:scRNA}, we use \href{https://openproblems.bio/}{OpenProblems.bio}'s default scVI settings for the encoder, decoder, latent dimension, likelihood, and optimization schedule.
They recommend training on all the data without a separate validation set since they compute scIB metrics on the full dataset after 400 epochs of training.
We, however, perform an 80/10/10 split into training, validation, and test sets in order to compute realistic estimates of the log marginal likelihood.

Our code is freely available at \url{https://github.com/astirn/VampPrior-Unleashed}.
We used a variety of compute environments for our experiments.
Most models and datasets can run on a modern laptop.
For bulk experiment runs, we ran multiple training processes in parallel on an Nvidia L40 GPU.

\subsection{Metrics}
\label{supp:experimental-setup:metrics}

\Cref{sec:experiments:unleashing,sec:experiments:clustering} report four test-set quantities.
Our primary measure of model evidence is the per-datum log marginal likelihood, estimated with an IWAE bound \parencite{burdaImportanceWeightedAutoencoders2016}.
We decompose the ELBO into its \emph{distortion} and \emph{rate} \parencite{alemiFixingBrokenELBO2018} to expose \emph{how} a prior buys its evidence: distortion measures reconstruction and rate measures $\E_{x}\KL(q_{\phi}(Z|X=x)\|p(Z))$, the cost in nats of coding the posterior under the prior.
We also compute the marginal KL \parencite{hoffmanElboSurgeryAnother2016}, which measures the divergence between the aggregated posterior and the prior.
Finally, we report the NMI between each model's MAP cluster assignment $\arg\max_j q(C_i=j|X_i=x_i)$ and the held-out class labels.

\subsubsection{Log Marginal Likelihood}
\label{supp:experimental-setup:metrics:evidence}

Let $\D = \{x_i\}_{i=1}^N$ denote the test dataset.
We consider models with both \emph{local} latent variables $Z_i$ (one per observation) and a \emph{global} latent variable $V$ shared across all observations.
For our DP variants, the global variable $V \in (0,1)^{K-1}$ are the stick-breaking proportions that determine the mixing proportions.
In all cases, we use $M=5{,}000$ samples and compute the per-datum log marginal likelihood using the checkpoint corresponding to the highest validation ELBO.

\paragraph{Standard IWAE (No Global Latent Variable)}

When the prior $p(Z)$ has no global parameters, the importance-weighted estimate of the log marginal likelihood is the standard IWAE bound \parencite{burdaImportanceWeightedAutoencoders2016}:
\begin{equation}
  \log p(X=x_i)
  \;\approx\;
  \log \frac{1}{M} \sum_{k=1}^{M}
  \frac{p_{\theta}(X=x_i \mid Z=z_k)\, p(Z=z_k)}{q_{\phi}(Z=z_k \mid X=x_i)},
  \qquad z_k \sim q_{\phi}(Z \mid X=x_i).
  \label{eq:iwae}
\end{equation}
Averaging \cref{eq:iwae} for $i\in[N]$ gives the per-datum log marginal estimate.

\paragraph{Two-level Importance Sampling with a Global Latent Variable.}

When the prior factorizes as $p(Z,V) = p(Z\mid V)\,p(V)$, the dataset marginal is
\begin{equation}
  \log p(\D)
  = \log \int p(V=v)\prod_{i=1}^{N}
  \int p_{\theta}(X=x_i \mid Z=z_i)\,p(Z=z_i\mid V=v)\,\mathrm{d}z_i\;\mathrm{d}v.
  \label{eq:joint-marginal}
\end{equation}
We estimate \cref{eq:joint-marginal} with a nested importance sampler using the variational posterior $q(V)$ as the proposal distribution for the outer integral and $q_{\phi}(Z\mid X=x_i)$ for each inner integral.
We first draw $M$ global samples $v^{(m)} \sim q(V)$ and compute log importance weights
\begin{equation}
  \ell_m^{\mathrm{global}}
  = \log p(V=v^{(m)}) - \log q(V=v^{(m)}).
  \label{eq:global-weights}
\end{equation}
Then, for each $v^{(m)}$ and each datapoint $x_i$, we draw $M$ local samples $z_k \sim q_{\phi}(Z \mid X=x_i)$ and form the inner log importance weight sum
\begin{equation}
  \ell_{i,m}^{\mathrm{local}}
  = \log \frac{1}{M} \sum_{k=1}^{M}
  \frac{p_{\theta}(X=x_i \mid Z=z_k)\,p\!\left(Z=z_k \mid V=v^{(m)}\right)}{q_{\phi}(Z=z_k \mid X=x_i)}.
  \label{eq:local-weights}
\end{equation}
Because observations are conditionally independent given $V$, the joint inner estimate decomposes as $\sum_i \ell_{i,m}^{\mathrm{local}}$, so the per-datum sums are accumulated batch-by-batch for each outer sample $v^{(m)}$.
The dataset's log marginal is then estimated as
\begin{equation}
  \log p(\D)
  \;\approx\;
  \log \frac{1}{M}
  \sum_{m=1}^{M}
  \exp\!\left(
  \ell_m^{\mathrm{global}}
  + \sum_{i=1}^{N} \ell_{i,m}^{\mathrm{local}}
  \right),
  \label{eq:two-level-iwae}
\end{equation}
which we divide by $N$ to obtain the per-datum log marginal estimate.

\subsubsection{Normalized Mutual Information}
\label{supp:experimental-setup:metrics:nmi}

We deliberately choose NMI over ARI as our clustering criterion.
ARI is built from pairwise co-assignment agreements, so splitting a true class into two coherent sub-clusters breaks a quadratic number of within-class pairs and is penalized almost as harshly as merging two distinct classes.
NMI's information-theoretic framing instead asks how much of the label variable a partition determines, and a refinement of the true partition retains that information.
Since no prior herein is told the number of classes, a criterion should not conflate refinement with error, making NMI more appropriate than ARI in our setting.
A sparsity-inducing prior should be free to discover a semantically meaningful substructure (e.g., sub-clustering a FashionMNIST \textsc{shirt} by sleeve length or a cell-type by disease state).
Our choice is also conservative with respect to our claims: NMI punishes the \textbf{VampPrior} less harshly than ARI when it over-clusters.

In \cref{sec:experiments:unleashing,sec:experiments:clustering,sec:experiments:pareto}, we compute NMI on the full dataset set using the checkpoint corresponding to the highest validation NMI.
In \cref{sec:experiments:scRNA}, we compute NMI on the full dataset using the final checkpoint to mirror the evaluation protocol \href{https://openproblems.bio/}{OpenProblems.bio} uses for scIB metrics.

\subsubsection{scIB Metrics}
\label{supp:experimental-setup:metrics:scib}

The single-cell Integration Benchmark (scIB) \parencite{lueckenBenchmarkingAtlaslevelData2022} evaluates scRNA-seq integration methods along two complementary axes---biological conservation and batch correction---and combines them into an aggregate total.
Each metric is rescaled to $[0,1]$ so that higher values are better.
The aggregate score for each axis is the unweighted mean of its constituent metrics, and the total score is the weighted mean of the two axes: $0.6$ for biological conservation and $0.4$ for batch correction.
Following \href{https://openproblems.bio/}{OpenProblems.bio}, we compute these metrics on the full dataset rather than a held-out subset, using the final model checkpoint.
In the case of scVI, an embedding-based method, these metrics apply to the latent embedding rather than the predicted gene expression.

We chose scIB's $k$-means option for ARI and NMI over Leiden clustering.
Using Leiden for clustering-based scores requires sweeping resolutions and picking the best match to ground truth, which can introduce bias and is computationally expensive.
Furthermore, Leiden clustering has notorious run-to-run variability, making it less reliable for consistent evaluation especially with our limited number of runs (three per configuration due to computational constraints).
In contrast, $k$-means clustering fixes the number of clusters in advance to the number of ground-truth cell types, which avoids the need for resolution sweeps and reduces computational overhead.

We omit the principal component regression (PCR) comparison, a default batch-correction scIB metric.
PCR comparison scores the reduction in variance explained by batch and cannot distinguish removing batch effects from removing all structure, and therefore can reward discarding biological variation.
Indeed, an embedding drawn from a standard normal attains a near-optimal PCR score, since an embedding with no gene-expression information cannot explain any batch variance---that is scVI with a standard normal prior that suffers posterior collapse trivially maximizes PCR.
We found that PCR scores varied so widely across methods that batch-correction ranking was largely dominated by PCR.

Several of the scIB metrics below compute an average silhouette width (ASW).
ASW measures the separation of clusters, where $-1$ indicates cluster overlap and $1$ indicates well separated clusters.
Given $a_i$ as the average distance between cell $i$ and all other cells in the same cluster and $b_i$ as the minimum average distance between cell $i$ and all cells in the \emph{nearest} other cluster, the silhouette width for cell $i$ is defined as
\begin{align}
  s(i) = (b_i - a_i) / \max(a_i, b_i) \label{eq:silhouette-width}
\end{align}
The ASW is the mean of $s(i)$ over all cells.

\paragraph{Biological Conservation}

\begin{itemize}[leftmargin=*,nosep]
  \item \textbf{Isolated labels} identifies cell types present in the fewest batches and measure whether these rare cell types remain distinct from other cell types after integration using ASW. This metric penalizes merging rare cell types into clusters of more populous cell types. The ASW is rescaled to $[0,1]$ via $(\text{ASW} + 1)/2$ and averaged over the set of isolated labels.
  \item \textbf{KMeans ARI} is the ARI between $k$-means cluster assignments of the latent embeddings and the ground-truth cell-type labels. The scIB package sets $k$ equal to the number of ground-truth cell types.
  \item \textbf{KMeans NMI} is the NMI between the same $k$-means partition and the cell-type labels. The scIB package sets $k$ equal to the number of ground-truth cell types.
  \item \textbf{Silhouette label} is ASW with respect to cell-type labels in the latent space. ASW is rescaled from $[-1,1]$ to $[0,1]$ via $(\text{ASW} + 1)/2$. Higher values indicate that cells lie closer to cells of their own type than to cells of the nearest other type.
  \item \textbf{cLISI} stands for \underline{c}ell-type \underline{L}ocal \underline{I}nverse \underline{S}impson's \underline{I}ndex. To compute cLISI, scIB first finds the $k$-nearest neighbors of each cell. Then from each cell's neighbor list, the inverse Simpson's index is calculated to measure the diversity of cell-type labels within the neighborhood. Then, cLISI is rescaled as $(L - \mathrm{cLISI})/(L - 1)$, where $L$ is the number of cell-type labels, so that $1$ indicates perfectly pure neighborhoods.
\end{itemize}

\paragraph{Batch Correction}

\begin{itemize}[leftmargin=*,nosep]
  \item \textbf{BRAS} stands for \underline{B}atch \underline{R}emoval \underline{A}dapted \underline{S}ilhouette \parencite{rautenstrauchShortcomingsSilhouetteSinglecell2026} and replaces the batch ASW in the version of the scIB package we use (0.5.9). Whereas standard batch ASW measures a cell's separation from the nearest other batch only, BRAS uses the mean distance to cells of all other batches in lieu of $b_i$ in \cref{eq:silhouette-width}. Per-cell scores are normalized to $[0,1]$, averaged within each cell type, and then averaged across cell types.
  \item \textbf{Graph connectivity} measures how well the $k$-nearest neighbor graph connects cells of the same type. In particular, it normalizes the size of the largest connected component within the subgraph of each cell type by the total number of cells of that type. This score is then averaged across all cell types. High values indicate that cells of the same type are connected across batches rather than split into batch-specific fragments.
  \item \textbf{kBET} is the $k$-Nearest Neighbor Batch Effect Test \parencite{buttnerTestMetricAssessing2019a} and compares the label composition of each cell's neighborhood to the global composition. For each cell type, kBET rewards a batch (i.e., donor) composition that matches the global composition--that is cell-type neighborhoods should have samples from all batches.
  \item \textbf{iLISI} stands for \underline{i}ntegration Local \underline{I}nverse \underline{S}impson's \underline{I}ndex is the negative analog of cLISI and is with respect to batch labels rather than cell-type labels: $(\mathrm{iLISI} - 1)/(B - 1)$.
\end{itemize}

\subsection{Pseudo-input Initialization}
\label{supp:experimental-setup:initialization}

Because the \textbf{VampPrior}, \textbf{VMM}, and our proposed priors update pseudo-inputs $u$ via gradient ascent, their initialization is itself an important hyperparameter.
We consider three strategies.
First and most familiar, \texttt{data} initialization draws $u$ uniformly without replacement from the training set as \textcite{tomczakVAEVampPrior2018} prescribe.
Second, \texttt{random} ignores the data and initializes $u\sim\N(0,I)$.
Lastly, our \texttt{stratified} initialization performs stratified sampling of the data over class labels and then interleaves pseudo-input initializations according to the class label.
This strategy ensures each class is represented in the first number-of-class-labels pseudo-inputs.
In the face of a DP's exponential decaying mixture proportions, this strategy lowers the risk of redundant pseudo-inputs starving later components.
From our perspective, this strategy is less about peeking at labels a priori and more about reducing seed-to-seed variance in the latent structure discovered.
One could easily substitute pre-training clustering (e.g., k-means or Leiden clustering) assignments for the class labels in the stratified initialization.
Another extension we did not explore is to interleave pseudo-input initializations that respect label proportions.

\Cref{sec:experiments:unleashing,sec:experiments:clustering} consider all three initialization strategies.
\Cref{fig:sweep_k:fashion_mnist:data,fig:sweep_k:fashion_mnist,fig:sweep_k:mnist,fig:sweep_alpha:fashion_mnist,fig:sweep_alpha:mnist} all specify which initialization strategy produced the result.
\Cref{sec:experiments:pareto} aggregates performance across the \texttt{data} and \texttt{random} strategies for the \textbf{VampPrior}, the \textbf{VMM}, and \textbf{VampPrior-Unleashed} and \texttt{stratified} and \texttt{random} for the DP variants (\textbf{DP-VampPrior} and \textbf{DP-VampPrior-Unleashed}).
While \texttt{stratified} initialization peeks at the labels, it helps mitigate the risk of redundant pseudo-inputs dominating the mixture.
In the event that class labels are not feasible other pre-clustering algorithms may be employed (e.g., k-means or Leiden clustering).
We observe that \texttt{random} generally offered the highest per-datum log marginal, but had lower NMI---although \texttt{random} peak NMI across seeds often was in the inter-quartile range of the \texttt{data} and \texttt{stratified} strategies.
As expected, models that have uniform mixture proportions--\textbf{VampPrior}, \textbf{VMM}, and \textbf{VampPrior-Unleashed}--had no theoretical benefit from \texttt{stratified} initialization in \cref{sec:experiments:unleashing,sec:experiments:clustering}.

As the main text mentions, it is quite common that scRNA-seq datasets provide coarse cell-type labels.
We use \texttt{data} initialization for \textbf{VMM} as they prescribe and \texttt{stratified} initialization for our \textbf{DP-VampPrior-Unleashed}.

\subsection{Hyperparameter Grids}
\label{supp:experimental-setup:hyperparameters}

\Cref{sec:experiments:unleashing} sweeps the number of components $K\in\{1,10,20,40,60,80,100\}$ over $20$ seeds for the \textbf{VampPrior} and \textbf{VampPrior-Unleashed} for all three of our pseudo-input initialization strategies (\cref{supp:experimental-setup:initialization}).
\Cref{sec:experiments:clustering} sweeps the DP's concentration parameter $\alpha\in\{0.1, 0.5, 0.75, 1.0, 2.0, 4.0, 10.0\}$ over $20$ seeds for the \textbf{DP-VampPrior} and \textbf{DP-VampPrior-Unleashed} for all three of our pseudo-input initialization strategies (\cref{supp:experimental-setup:initialization}).
The variational stick-breaking distributions initialize to match the prior (i.e., $\text{Beta}(1, \alpha)$.

\Cref{sec:experiments:pareto} reuses these parameter grids, but expands the \textbf{DP-VampPrior} and \textbf{DP-VampPrior-Unleashed} parameter space.
This expansion supplements the original grid with $\alpha\in\{11.0,100.0\}$ and, for $\alpha=11.0$, initializes the variational stick-breaking distributions to begin with uniform mixture proportions.
These additional values allow to the DP to explore solutions closer to the $K=100$ uniform mixture proportions used by the \textbf{VampPrior} and \textbf{VampPrior-Unleashed} and thereby approximate the log marginal's upper bound for the DP variants.
Following \textcite{stirnVampPriorMixtureModel2025}, we adjust the prior's strength by setting a pseudo dataset size for $N\in\{64, 256, 1024, 4096, 16384, 60000\}$; this has the effect of up-weighting the prior's strength as the ratio of the pseudo dataset size to the actual dataset size.
They suggest this method as an alternative to treating batch size as the dataset size when its inconvenient to do so.
To limit the parameter space and run-to-run variance, we use \texttt{data} and \texttt{random} initialization strategies for all non-DP variants (\textbf{VampPrior}, \textbf{VMM}, and \textbf{VampPrior-Unleashed}) and \texttt{stratified} and \texttt{random} for the DP variants (\textbf{DP-VampPrior} and \textbf{DP-VampPrior-Unleashed}).
For reasons \cref{supp:experimental-setup:initialization} discusses, we think this was fair choice.

\section{Additional Hyperparameter Sweeps}
\label{supp:sweeps}

The figures below contain the additional sweeps referenced in \cref{sec:experiments:unleashing,sec:experiments:clustering}.
All figures share the conventions of \cref{fig:sweep_k:fashion_mnist:data}.
Because each pair of variants share the hyperparameter settings and train on matched seeds, we pair the runs by seed, form the per-seed difference, and apply a one-sided Wilcoxon signed-rank test whose alternative is oriented by the metric (greater for log marginal and NMI, less for distortion, rate and marginal KL).
Differences with $p<0.05$ are annotated in the sweep figures by a dashed connector between the two conditional means, with triangles pointing in the direction of improvement for that metric (up for log marginal and NMI; down for distortion, rate and marginal KL).
We report these tests unadjusted for multiplicity: each is a separate question about a separate model configuration rather than a family of tests supporting a single claim.
Our conclusions rest on the consistency of the sign of the effect across the whole sweep rather than on any individual rejection.

\subsection{Varying Number of VampPrior Components}
\label{supp:sweep_k}

\begin{figure}[p]
  \centering
  \begin{subfigure}[t]{\textwidth}
    \centering
    \includegraphics[width=\textwidth]{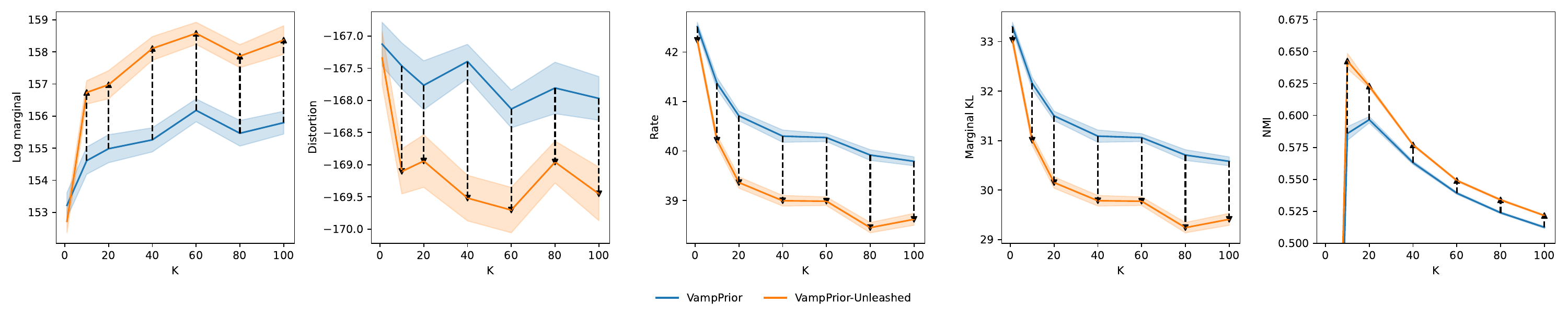}
    \caption{$u$ initialized from class-stratified training data}
    \label{fig:sweep_k:fashion_mnist:stratified}
  \end{subfigure}
  \begin{subfigure}[t]{\textwidth}
    \centering
    \includegraphics[width=\textwidth]{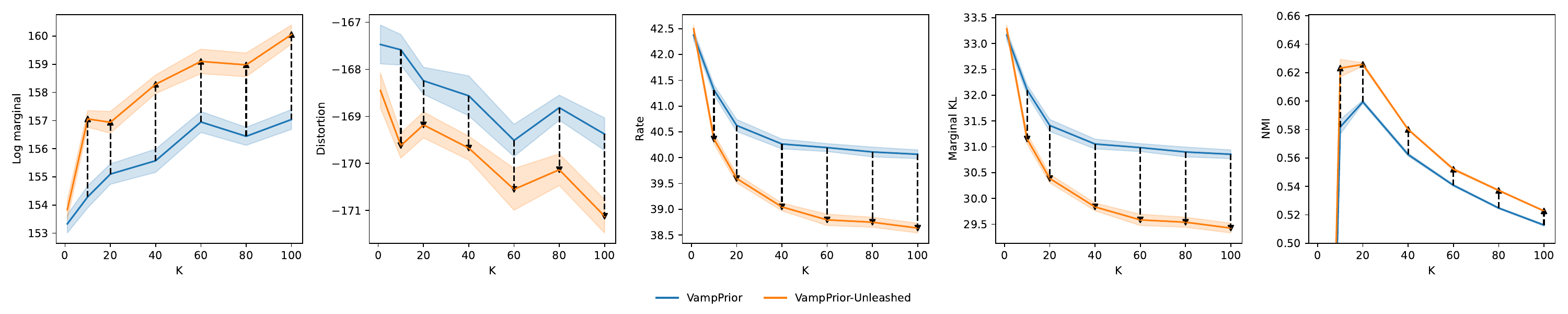}
    \caption{$u$ initialized from $\N(0,I)$}
    \label{fig:sweep_k:fashion_mnist:random}
  \end{subfigure}
  \caption{Effect of varying the number of components $K$ on FashionMNIST. \sweepAnnotations}
  \label{fig:sweep_k:fashion_mnist}
\end{figure}

\begin{figure}[p]
  \centering
  \begin{subfigure}[t]{\textwidth}
    \centering
    \includegraphics[width=\textwidth]{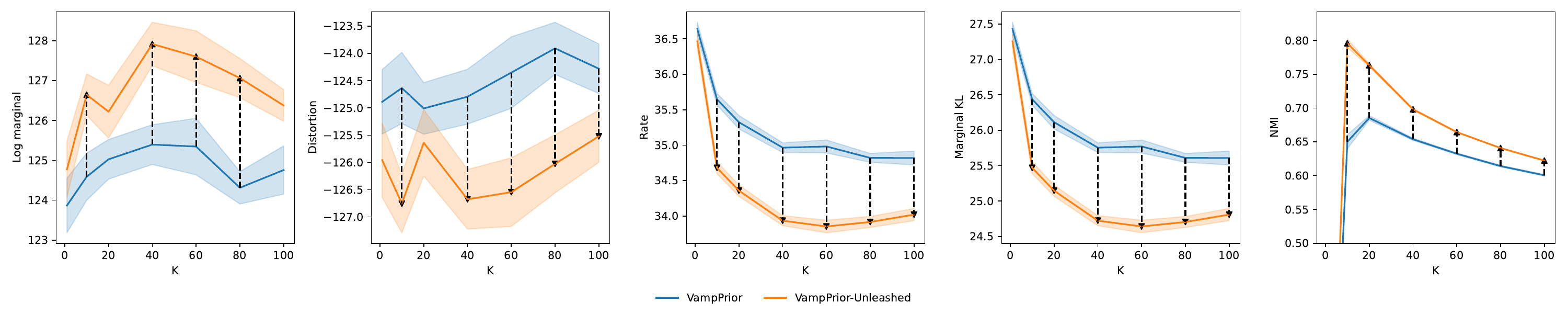}
    \caption{$u$ initialized from randomly selected training data}
    \label{fig:sweep_k:mnist:data}
  \end{subfigure}
  \begin{subfigure}[t]{\textwidth}
    \centering
    \includegraphics[width=\textwidth]{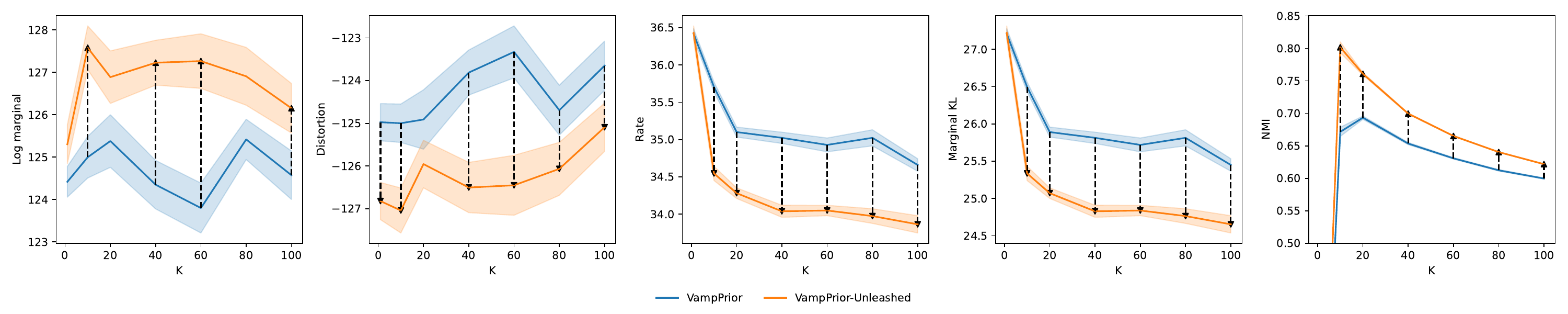}
    \caption{$u$ initialized from class-stratified training data}
    \label{fig:sweep_k:mnist:stratified}
  \end{subfigure}
  \begin{subfigure}[t]{\textwidth}
    \centering
    \includegraphics[width=\textwidth]{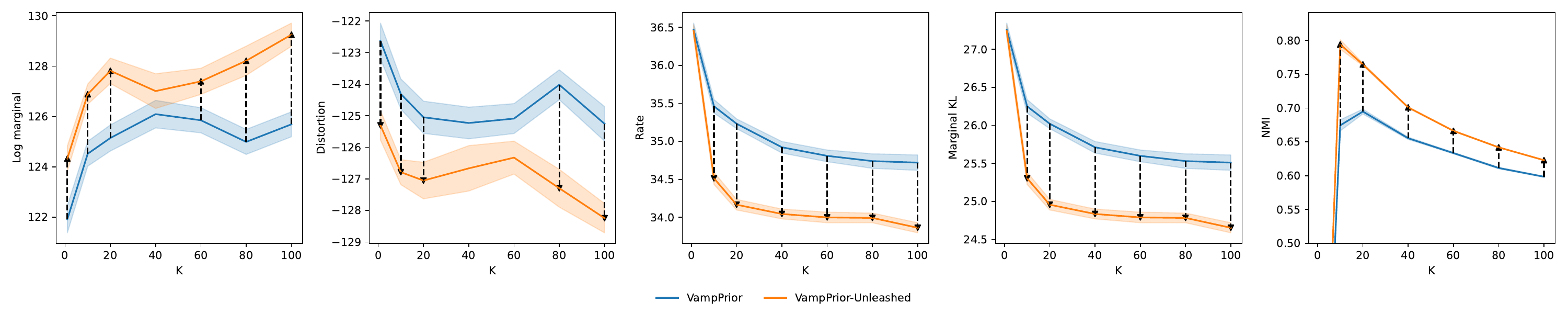}
    \caption{$u$ initialized from $\N(0,I)$}
    \label{fig:sweep_k:mnist:random}
  \end{subfigure}
  \caption{Effect of varying the number of components $K$ on MNIST. \sweepAnnotations}
  \label{fig:sweep_k:mnist}
\end{figure}

\Cref{sec:experiments:unleashing} only contained \cref{fig:sweep_k:fashion_mnist:data}, which sweeps $K$ over FashionMNIST with pseudo-inputs $u$ initialized using the \texttt{data} strategy.
\Cref{fig:sweep_k:fashion_mnist} contains the sweep for the \texttt{stratified} (\cref{fig:sweep_k:fashion_mnist:stratified}) and \texttt{random} (\cref{fig:sweep_k:fashion_mnist:random}) initialization strategies.
\Cref{fig:sweep_k:mnist} contains sweeps for all three initialization strategies on the MNIST dataset.

\subsection{Varying VampPrior DP Concentration}
\label{supp:sweep_alpha}

\begin{figure}[p]
  \centering
  \begin{subfigure}[t]{\textwidth}
    \centering
    \includegraphics[width=\textwidth]{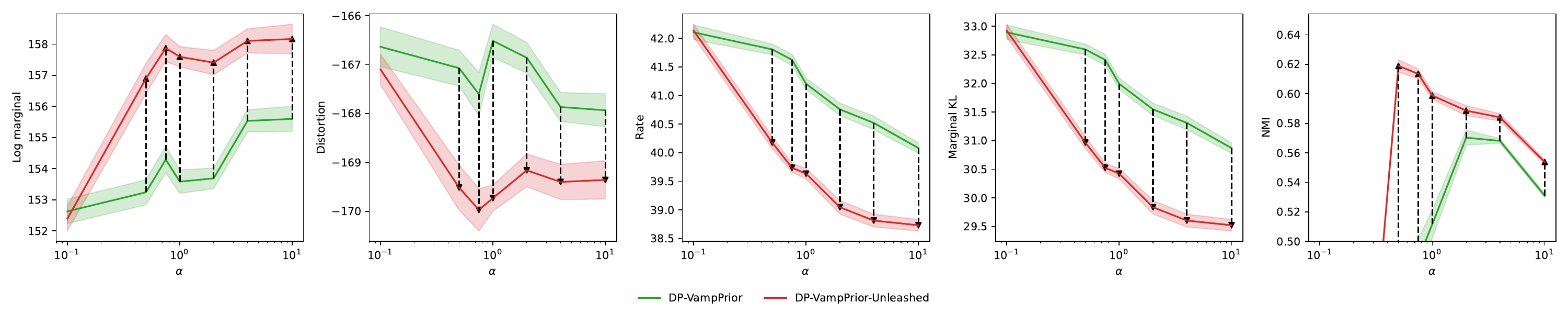}
    \caption{$u$ initialized from randomly selected training data}
    \label{fig:sweep_alpha:fashion_mnist:data}
  \end{subfigure}
  \begin{subfigure}[t]{\textwidth}
    \centering
    \includegraphics[width=\textwidth]{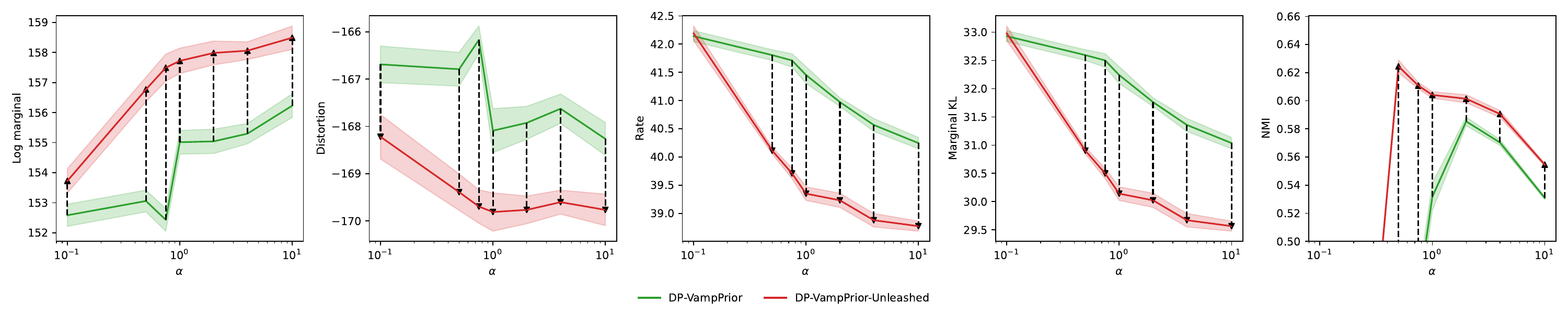}
    \caption{$u$ initialized from class-stratified training data}
    \label{fig:sweep_alpha:fashion_mnist:stratified}
  \end{subfigure}
  \begin{subfigure}[t]{\textwidth}
    \centering
    \includegraphics[width=\textwidth]{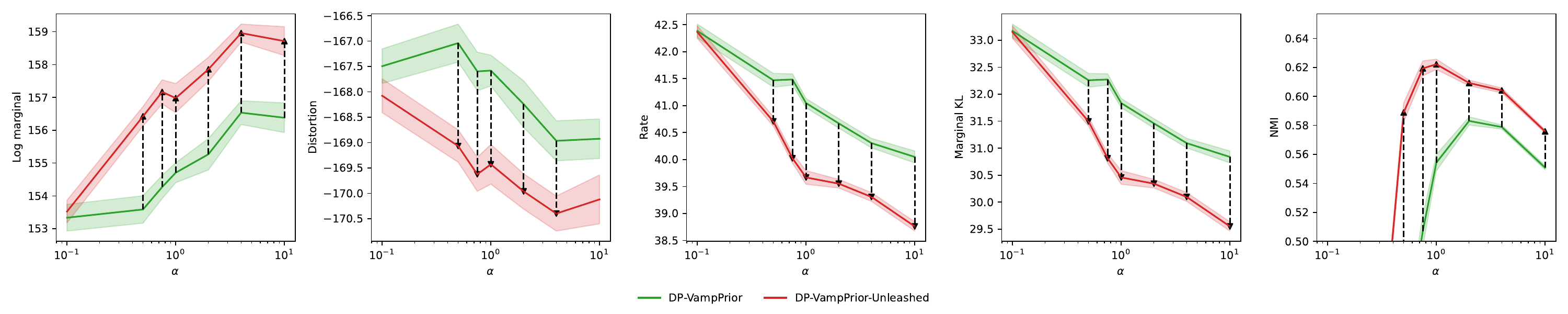}
    \caption{$u$ initialized from $\N(0,I)$}
    \label{fig:sweep_alpha:fashion_mnist:random}
  \end{subfigure}
  \caption{Effect of varying the concentration parameter $\alpha$ on FashionMNIST with truncation set to $K=100$. \sweepAnnotations}
  \label{fig:sweep_alpha:fashion_mnist}
\end{figure}

\begin{figure}[p]
  \centering
  \begin{subfigure}[t]{\textwidth}
    \centering
    \includegraphics[width=\textwidth]{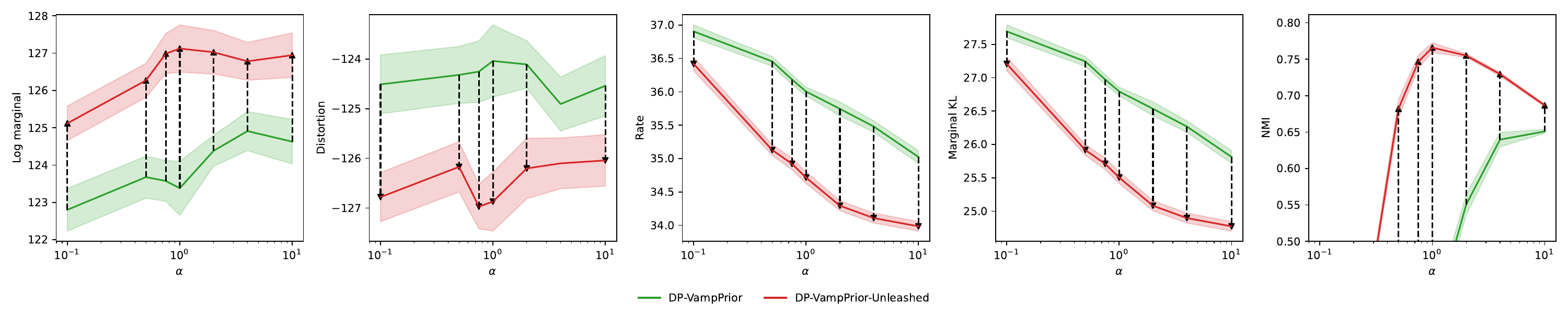}
    \caption{$u$ initialized from randomly selected training data}
    \label{fig:sweep_alpha:mnist:data}
  \end{subfigure}
  \begin{subfigure}[t]{\textwidth}
    \centering
    \includegraphics[width=\textwidth]{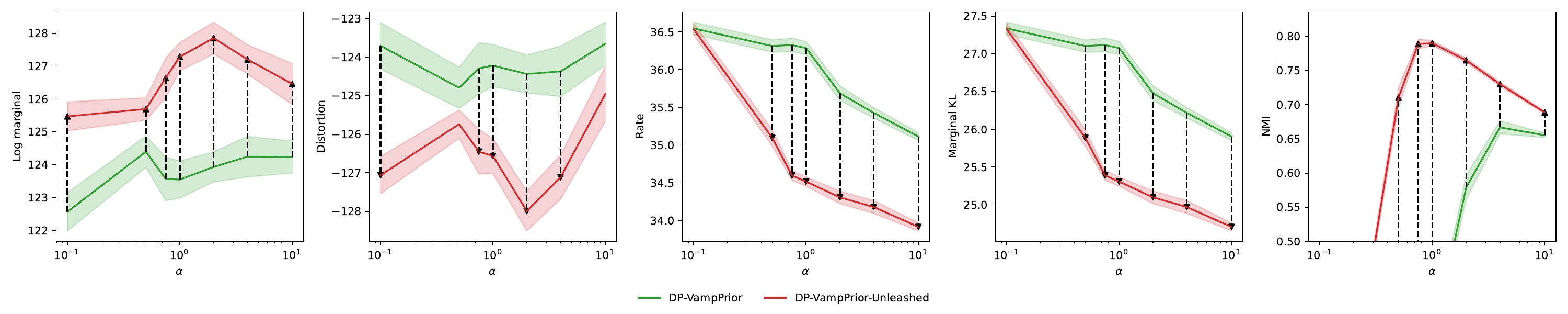}
    \caption{$u$ initialized from class-stratified training data}
    \label{fig:sweep_alpha:mnist:stratified}
  \end{subfigure}
  \begin{subfigure}[t]{\textwidth}
    \centering
    \includegraphics[width=\textwidth]{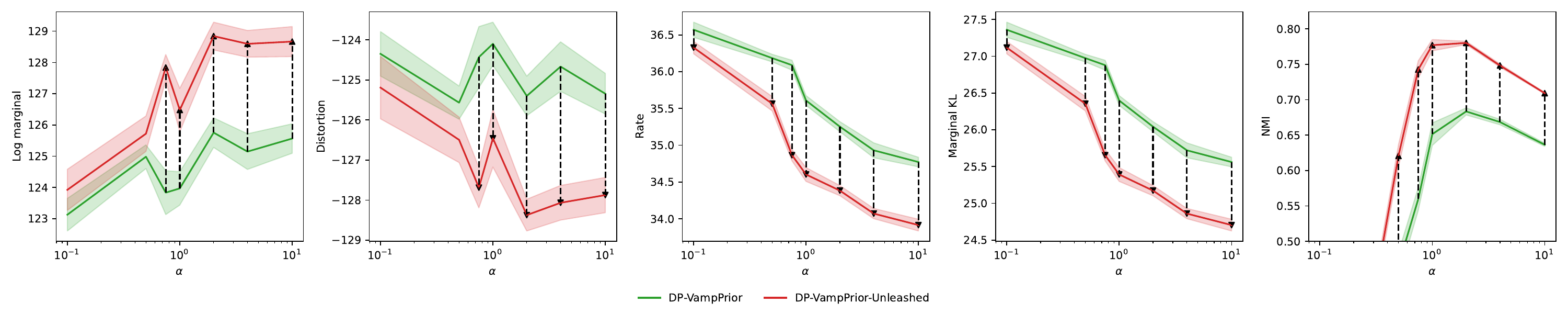}
    \caption{$u$ initialized from $\N(0,I)$}
    \label{fig:sweep_alpha:mnist:random}
  \end{subfigure}
  \caption{Effect of varying the concentration parameter $\alpha$ on MNIST with truncation set to $K=100$. \sweepAnnotations}
  \label{fig:sweep_alpha:mnist}
\end{figure}

\Cref{fig:sweep_alpha:fashion_mnist,fig:sweep_alpha:mnist} show the effect of varying the DP's concentration parameter $\alpha$ on FashionMNIST and MNIST, respectively.
Each plot has three subplots corresponding to the different initialization strategies: \texttt{data}, \texttt{stratified}, and \texttt{random}.

\clearpage

\clearpage

\clearpage

\clearpage

\clearpage

\section{Pairwise Pareto Comparisons}
\label{supp:pareto-diffs}

We utilize a Pareto analysis of the evidence-NMI curve in \cref{sec:experiments:pareto} because the priors do not always share the same hyperparameter axis.
Due to the intricacies of this analysis, however, we use a nonparametric bootstrap to quantify uncertainty.
Using $10{,}000$ bootstrap replicates, we resample the 20 seeds per configuration for every prior with replacement.
Specifically, there are 20 seeds per (prior, $u$-initialization, hyperparameter) tuple, where the hyperparameter axis is $K$ for the VampPrior and VampPrior-Unleashed, $\alpha$ for DP-VampPrior and DP-VampPrior-Unleashed, and pseudo dataset size for the VMM.
Under this ressampling, we recompute the entire downstream pipeline: the normalization bounds, the Pareto frontier, and every summary statistic derived therefrom.
Bootstrapping the entire analysis pipeline is essential here, because the frontier statistics are extremum-like functionals of the per-configuration means and their sampling distributions are neither symmetric nor centered on the point estimate.
\Cref{tab:pareto_means} therefore reports the point estimate alongside its $[2.5,97.5]$ percentile interval.
Each of the $10{,}000$ replicates share a random state and resample the same 20 seeds per configuration for all (prior, $u$-initialization, hyperparameter) tuples; this allows us to perform paired comparisons of the pipeline outputs across priors and compute the bootstrap probability $P(\Delta>0)$ that one prior exceeds another (i.e., $P(\text{prior A's metric} - \text{prior B's metric} > 0)$) in \cref{tab:pareto_diffs}.
In \cref{tab:pareto_means} we embolden the best prior per metric together with every prior we cannot distinguish from it (i.e.\ every prior whose paired comparison against the best satisfies $0.025<P(\Delta>0)<0.975$).
\Cref{tab:pareto_diffs} shows the paired bootstrap probabilities for all head-to-head comparisons.

\begin{table}[H]
  \caption{Paired bootstrap probabilities $P(\Delta>0)$ that the first prior of each comparison exceeds the second on every Pareto summary statistic reported in \cref{sec:experiments:pareto}. For $10{,}000$ replicates with shared seeds for every (prior, $u$-initialization, hyperparameter) tuple, we recompute the entire frontier pipeline to construct $P(\Delta>0)$. Values below $0.025$ or above $0.975$ indicate a difference we can resolve; note that for the distance to the ideal, $\min||P^{\ast}-(1,1)||_2$, smaller is better and the direction of the inequality is therefore reversed. \Cref{tab:pareto_means} uses these probabilities to determine its boldface.}
  \label{tab:pareto_diffs}
  \begin{center}
    \adjustbox{width=\textwidth}{\begin{tabular}{llllllll}
\toprule
 &  & $P(\Delta$ Hypervolume $> 0)$ & $P(\Delta$ Min $||P^{\ast} - (1,1)||_2$ $> 0)$ & $P(\Delta$ $P^{\ast}$ Log marginal $> 0)$ & $P(\Delta$ $P^{\ast}$ NMI $> 0)$ & $P(\Delta$ Max Log marginal $> 0)$ & $P(\Delta$ Max NMI $> 0)$ \\
Dataset & Comparison &  &  &  &  &  &  \\
\midrule
\multirow[t]{10}{*}{MNIST} & VampPrior $-$ VampPrior-Unleashed & 0.000 & 1.000 & 0.000 & 0.531 & 0.000 & 0.000 \\
 & VampPrior $-$ DP-VampPrior & 0.695 & 0.315 & 0.698 & 0.423 & 0.699 & 0.827 \\
 & VampPrior $-$ DP-VampPrior-Unleashed & 0.000 & 1.000 & 0.000 & 0.034 & 0.000 & 0.000 \\
 & VampPrior $-$ VMM & 0.000 & 1.000 & 0.001 & 0.003 & 0.001 & 0.000 \\
 & VampPrior-Unleashed $-$ DP-VampPrior & 1.000 & 0.000 & 1.000 & 0.537 & 1.000 & 1.000 \\
 & VampPrior-Unleashed $-$ DP-VampPrior-Unleashed & 0.272 & 0.970 & 0.471 & 0.064 & 0.433 & 0.782 \\
 & VampPrior-Unleashed $-$ VMM & 0.965 & 0.121 & 0.961 & 0.134 & 0.990 & 0.000 \\
 & DP-VampPrior $-$ DP-VampPrior-Unleashed & 0.000 & 1.000 & 0.000 & 0.027 & 0.000 & 0.000 \\
 & DP-VampPrior $-$ VMM & 0.000 & 1.000 & 0.000 & 0.001 & 0.000 & 0.000 \\
 & DP-VampPrior-Unleashed $-$ VMM & 0.994 & 0.006 & 0.989 & 0.560 & 0.997 & 0.000 \\
\cline{1-8}
\multirow[t]{10}{*}{FashionMNIST} & VampPrior $-$ VampPrior-Unleashed & 0.000 & 1.000 & 0.000 & 0.359 & 0.000 & 0.000 \\
 & VampPrior $-$ DP-VampPrior & 0.720 & 0.271 & 0.763 & 0.266 & 0.763 & 0.998 \\
 & VampPrior $-$ DP-VampPrior-Unleashed & 0.000 & 1.000 & 0.000 & 0.245 & 0.000 & 0.000 \\
 & VampPrior $-$ VMM & 0.000 & 1.000 & 0.000 & 0.459 & 0.000 & 0.000 \\
 & VampPrior-Unleashed $-$ DP-VampPrior & 1.000 & 0.000 & 1.000 & 0.304 & 1.000 & 1.000 \\
 & VampPrior-Unleashed $-$ DP-VampPrior-Unleashed & 0.725 & 0.613 & 0.755 & 0.131 & 0.788 & 0.836 \\
 & VampPrior-Unleashed $-$ VMM & 0.999 & 0.019 & 0.981 & 0.300 & 0.997 & 0.309 \\
 & DP-VampPrior $-$ DP-VampPrior-Unleashed & 0.000 & 1.000 & 0.000 & 0.271 & 0.000 & 0.000 \\
 & DP-VampPrior $-$ VMM & 0.000 & 1.000 & 0.000 & 0.475 & 0.000 & 0.000 \\
 & DP-VampPrior-Unleashed $-$ VMM & 0.993 & 0.018 & 0.904 & 0.629 & 0.955 & 0.027 \\
\cline{1-8}
\bottomrule
\end{tabular}
}
  \end{center}
\end{table}

\clearpage

\section{scRNA-seq Datasets}
\label{supp:scrna}

\Cref{tab:scrna_datasets} characterizes how strongly donor identity is confounded with biological composition in each atlas.
We quantify confoundment with Cramér's $V\in[0,1]$, which we compute from the $\chi^2$ test of independence on the contingency table of donor $\times$ cell-type labels. Here, $0$ indicates donors are unrelated to the biological labels and $1$ indicates donor identity fully determines them; the degrees of freedom are those of that test.
When $V$ is large, donors differ systematically in their cell-type composition, so batch-correction metrics that mix cells across batches, notably BRAS, can reward embeddings that blur biological structure; we therefore use $V$ only to contextualize the batch-correction columns of \cref{tab:scRNA}.

\begin{table}[H]
  \caption{Summary of single-cell RNA-seq datasets used in this study, including the number of cells, number of donors, Cramér's \(V\) statistic for the association between cell type and donor, \(k = \min(r, c) - 1\) for the \(r \times c\) donor-by-cell-type contingency table, and a qualitative assessment of association strength. Following \textcite{cohenStatisticalPowerAnalysis2013}, association is Low, Moderate, or High when \(V\) exceeds \(0.1/\sqrt{k}\), \(0.3/\sqrt{k}\), or \(0.5/\sqrt{k}\), respectively.}
  \label{tab:scrna_datasets}
  \centering
  \begin{tabular}{lrrrrl}
    \toprule
    Dataset              & \# Cells  & \# Donors & Cramér's \(V\) & Cohen's \(k\) & Association \\
    \midrule
    DKD                  & 39{,}176  & 11        & 0.145          & 10            & Moderate    \\
    Immune Cell Atlas    & 329{,}762 & 12        & 0.227          & 11            & High        \\
    GTEx v9              & 209{,}126 & 16        & 0.480          & 15            & High        \\
    Mouse Pancreas Atlas & 301{,}796 & 56        & 0.381          & 12            & High        \\
    HypoMap              & 384{,}925 & 24        & 0.055          & 12            & Low         \\
    Tabula Sapiens       & 483{,}152 & 29        & 0.444          & 28            & High        \\
    \bottomrule
  \end{tabular}
\end{table}

\end{document}